\documentclass{article}

\PassOptionsToPackage{numbers, compress}{natbib}

    \usepackage[final]{neurips_2025}

\usepackage{comment}
\usepackage[utf8]{inputenc} 
\usepackage[T1]{fontenc}    
\usepackage{hyperref}       
\usepackage{url}            
\usepackage{booktabs}       
\usepackage{amsfonts}       
\usepackage{nicefrac}       
\usepackage{microtype}      
\usepackage{xcolor}         
\usepackage{enumitem}
\usepackage{wrapfig}
\usepackage{lscape}

\usepackage{graphicx}
\usepackage{amsmath}
\usepackage{amsthm}
\newtheorem{definition}{Definition}
\DeclareMathOperator*{\argmin}{argmin}

\def\P{{\cal P}}

\usepackage{multirow}
\usepackage{adjustbox}
\usepackage{algorithm}
\usepackage{tcolorbox}
\usepackage{lmodern}
\usepackage{caption}

\usepackage{array}
\usepackage{multirow}
\usepackage{colortbl}
\usepackage{graphicx}

\usepackage{booktabs} 

\title{AutoOpt: A Dataset and a Unified Framework for Automating Optimization Problem Solving}

\author{
  Ankur Sinha, Shobhit Arora, and Dhaval Pujara \\
  Brij Disa Centre for Data Science and AI\\ 
  Indian Institute of Management Ahmedabad\\
  Ahmedabad, Gujarat 380015\\
  \texttt{\{asinha,shobhita,dhavalp\}@iima.ac.in}
}

\begin{document}

\maketitle

\begin{abstract}
This study presents \textit{AutoOpt-11k}, a unique image dataset of over 11,000 handwritten and printed mathematical optimization models corresponding to single-objective, multi-objective, multi-level, and stochastic optimization problems exhibiting various types of complexities such as non-linearity, non-convexity, non-differentiability, discontinuity, and high-dimensionality. The labels consist of the LaTeX representation for all the images and modeling language representation for a subset of images. The dataset is created by 25 experts following ethical data creation guidelines and verified in two-phases to avoid errors. Further, we develop \textit{AutoOpt} framework, a machine learning based automated approach for solving optimization problems, where the user just needs to provide an image of the formulation and \textit{AutoOpt} solves it efficiently without any further human intervention. \textit{AutoOpt} framework consists of three Modules: (i) M1 (\textit{Image\_to\_Text})- a deep learning model performs the Mathematical Expression Recognition (MER) task to generate the LaTeX code corresponding to the optimization formulation in image; (ii) M2 (\textit{Text\_to\_Text})- a small-scale fine-tuned LLM generates the PYOMO script (optimization modeling language) from LaTeX code; (iii) M3 (\textit{Optimization})- a Bilevel Optimization based Decomposition (BOBD) method solves the optimization formulation described in the PYOMO script. We use \textit{AutoOpt-11k} dataset for training and testing of deep learning models employed in \textit{AutoOpt}. The deep learning model for MER task (M1) outperforms ChatGPT, Gemini and Nougat on BLEU score metric. BOBD method (M3), which is a hybrid approach, yields better results on complex test problems compared to common approaches, like interior-point algorithm and genetic algorithm.


\end{abstract}

\section{Introduction}\label{sec:intro}
Optimization is an active field of research due to its potential to deliver substantial and sustainable benefits to users by providing efficient solutions to complex optimization problems that commonly arise in practice. 
There are efforts to generate mathematical programs from verbal explanations using the large-language models (LLMs) \cite{ahmaditeshnizi2024optimus,ma2024llamoco}, which currently works for simple problems and often requires iterations and further refinements to arrive at the right formulation. Little attention has been given to automate the solution of optimization problems stored in image-based formats, such as figures in research articles, scanned pages from books, handwritten notes, or whiteboard snapshots. Humans can easily read and interpret the information provided in images in the form of mathematical programming formulations, also known as mathematical models, mathematical programs, mathematical formulations, or simply optimization problems. However, this is not the case for machines as images lack the semantic structure \cite{schmitt2022accessible}, making it difficult for machines to understand the image content, i.e., poor machine readability. Therefore, such formulations always need to be represented in the form of a structured modeling language for the computer or an optimization solver to understand and solve it.

Very often in a classroom setting, research setting, or industrial setting, when a business or engineering problem is discussed, a mathematical formulation is worked out on the whiteboard, tablet, or paper and saved in the form of an image. After this, the tasks of converting the problem into a machine-readable format and solving it with a suitable solver, still remains. In most of the engineering and business schools, there are quite a few sessions devoted on solving such problems after the mathematical model is ready. Such tasks are usually mechanical and can be automated. In the current era of Artificial Intelligence (AI), humans aim to leverage machine learning by developing automated systems \cite{ivanov2023automated}, in which machines are primarily responsible for executing tasks with limited human intervention. 
In the context of the current study, it means that we want the machines to learn to interpret the mathematical formulations and participate in problem-solving tasks. 
However, to enable a machine to learn, it requires datasets that map mathematical formulations in images to their machine-readable representation in text format, which is currently lacking in the optimization community.
There exist studies, where researchers perform Optical Character Recognition (OCR) task \cite{blecher2023nougat}, in which the content within an image is recognized and converted into machine-encoded text, a machine-readable format that can be easily understood and processed by machines. Application of OCR to identify and convert text within images into machine-readable format is also known as text recognition \cite{chen2021text_recognition}, and in case of mathematical expressions, it is referred to as Mathematical Expression Recognition (MER) \cite{deng2017image,schmitt2024mathnet}. Optimization problems are commonly described through complex verbal explanations or large mathematical programs. When expressed as mathematical programs, it is not straightforward to apply existing OCRs and MERs to completely understand complex formulations and convert them into a machine-readable format. This study attempts to bridge this gap, which is of significant importance to the optimization community.


Tesseract OCR \cite{Tesseract_OCR_2007}, a well-known text recognition method, follows one-dimensional and line-by-line approach. It detects a single line of text from an image or PDF, sequentially identifies letters, words, or spacing in a considered line, and then moves to the next line. Such a one-dimensional or horizontal approach is not sufficient for MER, as mathematical expressions consist of several structural components such as subscripts, superscripts (exponents), fractions, matrices, etc. These components are characterized by the relative spatial positions of characters and symbols in expression, and they convey specific semantic relationships and mathematical meaning. To detect and preserve this meaning into machine-encoded text, methods need to analyze the content not only horizontally but also vertically, in a two-dimensional manner \cite{gervais2024mathwriting}. 
Further, MERs which work with single line mathematical expressions may also not suffice for mathematical programs, which are often multilined and contain interlinked information, such as variable(s), parameter(s), objective function(s) and constraint(s). Therefore, there is a need for MERs that have the ability to understand mathematical programs as a whole rather than in pieces.

\begin{figure*}[hbt]
\centering
\includegraphics[width=1\textwidth]{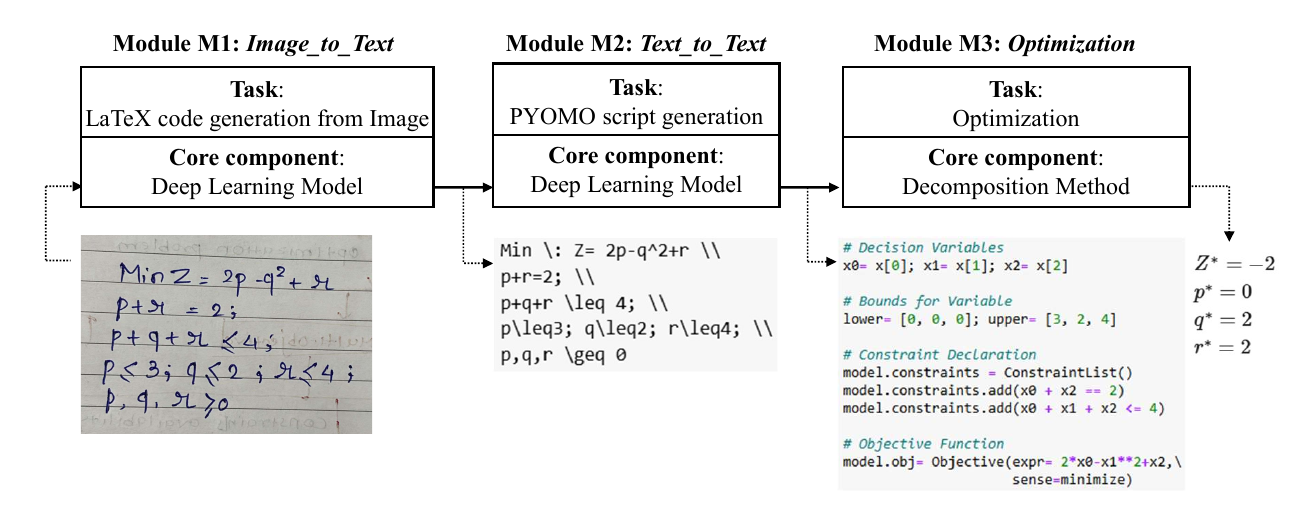}
\caption{\textit{AutoOpt} framework: workflow demonstration using an example}
\label{fig:AutoOpt}
\end{figure*}

In the case of optimization problems, after converting a mathematical model into machine-readable format (for example, LaTeX) using MER, there is a scope of further value addition by creating a setup that can extract the relevant data from the converted optimization problem and fit it into a predefined programming structure (for example, mathematical modeling languages, like AMPL, PYOMO, etc.). This program structure can be passed to a mathematical solver or an optimization technique implemented in the computational system to solve the respective optimization problem effectively. In this way, the task of solving optimization problems can be automated, where user only needs to provide an image of the mathematical program, and an efficient solution to the respective mathematical program can be retrieved with little human intervention. This study achieves the same by proposing \textit{AutoOpt}, an automated optimization framework. The dataset and the code associated with \textit{AutoOpt} framework is being made publicly available\footnote{The \textit{AutoOpt-11k} dataset can be accessed through the link \url{https://www.kaggle.com/datasets/ankurzing/autoopt-11k}, and the code for \textit{AutoOpt} framework can be accessed through the link \url{https://github.com/Shobhit1201/AutoOpt}.}. The workflow of the {\it AutoOpt} framework is provided in Figure~\ref{fig:AutoOpt}, which contains three modules described next:
\begin{itemize}[leftmargin=1em] 
\vspace{-2mm}
    \item \textbf{M1 (\textit{Image\_to\_Text})}: A deep learning module that takes an image as an input and generates LaTeX code corresponding to the mathematical model in image.\\
    {\it Contribution:} Releasing an Image to LaTeX dataset ({\it AutoOpt-11k}) with 11,554 mathematical programs that are a mix of handwritten and typeset (printed) images. A deep learning architecture is also proposed to capitalize on this dataset.
    \item \textbf{M2 (\textit{Text\_to\_Text})}: A deep learning model that takes LaTeX code from module M1 as input and generates a PYOMO script.\\
    {\it Contribution:} Releasing an Image/LaTeX to PYOMO dataset with 1,018 mathematical programs (a subset of {\it AutoOpt-11k}). A pre-trained deep learning model is fine-tuned to develop this module.
    \item \textbf{M3 (\textit{Optimization})}: An effective bilevel optimization based decomposition (BOBD) method, implemented in Python, solves the problem using the PYOMO script obtained from module M2.\\
    {\it Contribution:} We build up on a recently proposed approach \cite{sinha2024decomposition} by automating the decomposition task using machine learning.
\end{itemize}
\vspace{-2mm}
\vspace{-2mm}
Note that it is possible to generate the final PYOMO script directly from an image using a single deep learning model. However, we adopt a two-stage approach, first generating LaTeX and then converting it to PYOMO, for several practical reasons. A two-step design enhances the ease of verification, as the intermediate LaTeX output serves as a human-readable checkpoint to inspect the accuracy of the model interpretation prior to code generation. Moreover, not all generated scripts are guaranteed to be executable due to errors of incompleteness in mathematical model description. 

The deep learning models used in \textit{AutoOpt} framework (modules M1 \& M2) must be trained on a rich, diverse, well-curated, and representative dataset of mathematical models to ensure effectiveness of the proposed approach. The literature offers a wide range of image datasets designed to train AI models for MER \cite{deng2017image, gervais2024mathwriting, schmitt2024mathnet,xie2023icdar,aida_calculus_dataset}. These datasets contain images of single-lined and small mathematical expressions typically drawn from various mathematical streams such as algebra, calculus, geometry, probability, statistics, etc. Hence, from the perspective of mathematics, these datasets are quite general in nature. However, the objective of automating optimization requires a rich dataset containing images of mathematical programs corresponding to optimization problems with varying levels of complexity. To the best of our knowledge, such a dataset does not exist currently, i.e., a dataset specifically for the optimization domain. We bridge this gap by developing \textit{AutoOpt-11k} image dataset, a collection of more than 11,000 mathematical programs. \textit{AutoOpt-11k} covers small-to-large scale theoretical and real-life problems from various categories of optimization problems such as constrained, unconstrained, linear, non-linear, convex, non-convex, single-objective, multi-objectives, etc. For each image of a mathematical program, that can be handwritten, printed or a mix of handwritten and printed, we provide its LaTeX representation. We provide a PYOMO representation for a subset of these images. The dataset has been created with the support of 25 experts and has been verified in two-phases to avoid errors.


The paper is organized as follows: The dataset development procedure and characteristics of the \textit{AutoOpt-11k} dataset are provided in Section~\ref{sec:dataset_creation}. Mechanism for solving an optimization problem using the automated optimization framework, \textit{AutoOpt}, is demonstrated using an example along with a detailed description of the modules, M1, M2 and M3, in Section~\ref{sec:AutoOpt}. 
The experimental results are provided in Section~\ref{sec:AutoOpt} and also in the Appendices. Finally, concluding remarks, future research directions and limitations are discussed in Section~\ref{sec:conclusions}.

\section{AutoOpt-11k Dataset}\label{sec:dataset_creation}
\begin{figure*}[hbt]
\centering
\includegraphics[width=1.02\textwidth]{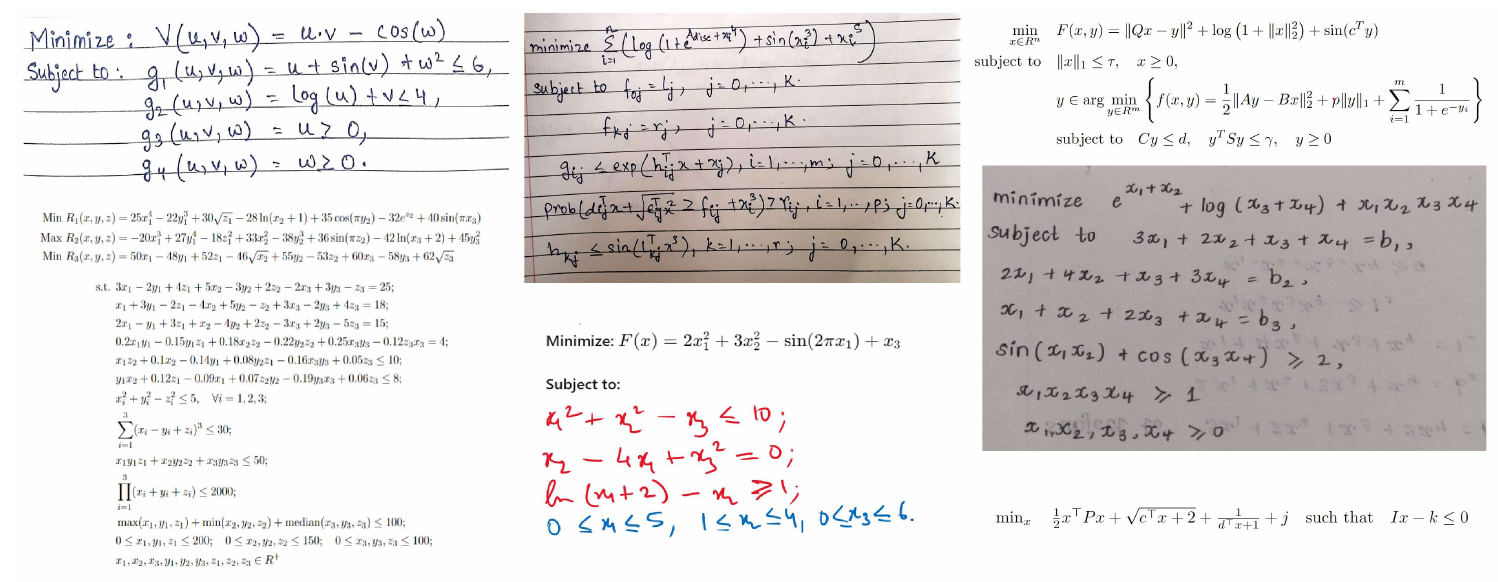}
\caption{A sample of 7 images from the {\it AutoOpt-11k} dataset.}
\label{fig:dataset}
\end{figure*}

\vspace{-2mm}
In this section, we discuss the key characteristics and the development process of \textit{AutoOpt-11k}, an image dataset of mathematical programs created in this study. 
\textit{AutoOpt-11k} consists of 11,554 images, each illustrating a distinct mathematical model corresponding to an optimization problem drawn from domains such as science, engineering, business, and related fields. Of these, 5,070 images feature handwritten mathematical models created manually by human annotators, while the remaining 6,484 images present typeset models taken from printed sources or generated using computer system. Apart from writing on paper, it is now common for people to write on tablets, electronic boards or touch screens, where typeset and handwritten text may appear together. The created data set incorporates these kinds of variations. A small sample of images that are part of the dataset is shown in Figure~\ref{fig:dataset}.

\vspace{-2mm}
The \textit{AutoOpt-11k} dataset has been systematically compiled to capture the broad diversity observed in optimization problems across theoretical and applied contexts. It includes single-objective optimization problems \cite{gill1981practical}, which involve the maximization or minimization of a single criterion, as well as multi-objective problems \cite{deb2001multi}, where multiple, often conflicting, objectives must be optimized simultaneously. The dataset also contains bilevel/multi-level optimization problems \cite{sinha2017review}, which are hierarchical in nature and involve nested decision-making structures, and stochastic problems \cite{birge2011introduction}, which involve uncertainty in the problem definition. From the perspective of constraints \cite{fletcher2000practical}, both constrained and unconstrained optimization problems are included, enabling the training of models that can interpret a wide range of structural conditions.

In mathematical modeling, optimization problems are typically described in either a general form, where parameters such as coefficients in the objective function and constraints are left symbolic, or in a numerical form, where all such values are explicitly provided. This distinction is crucial in both theoretical exposition and practical implementation. Accordingly, \textit{AutoOpt-11k} includes problem statements in both general and numerical forms. 
Further, mathematical models can be expressed either using compact matrix-vector notation or in a scalar format involving only scalar operations. Matrix-vector notation is often preferred in compact representations, especially in linear algebraic formulations, while algebraic expressions are commonly used in detailed, problem-specific contexts. \textit{AutoOpt-11k} includes mathematical programs expressed in both styles.

The functional form of the objective function and constraints contributes significantly to the complexity of an optimization problem. These functions may exhibit various mathematical properties, such as being linear or non-linear, convex or non-convex, continuous or discontinuous, and differentiable or non-differentiable. These characteristics directly influence the selection of appropriate solution methods and the computational difficulty of solving the problem. In addition, the dimensionality of the problem—defined by the number of decision variables and constraints—plays a key role in determining its scale and complexity. To ensure comprehensive coverage, \textit{AutoOpt-11k} incorporates problems spanning a wide range of functional behaviors and scales, thus including mathematical programs across a broad landscape of optimization scenarios. The diverse composition of \textit{AutoOpt-11k} dataset is provided in Table~\ref{tab:AutoOpt_11k} along with the count of images of each type. The optimization problems have been taken from or inspired by the sources mentioned in Table~\ref{tab:autoopt_sources}.

\begin{table}[t]
\centering
\caption{Composition of \textit{AutoOpt-11k} dataset based on the characteristics of optimization problems}
\label{tab:AutoOpt_11k}
\scalebox{0.78}{
\begin{tabular}{llrl}
\toprule
\textbf{} & \textbf{Type} & \textbf{Count} & \textbf{Description} \\
\midrule
\multirow{3}{*}{\centering {\it AutoOpt-11k}} 
  & Handwritten & 5,070 & Written by hand on paper, tablet, electronic book, etc. \\
  & Typeset     & 6,484 & Printed format extracted from books, articles, etc. \\
  & Total       & 11,554 & \\ 
\midrule
\multirow{4}{*}{\centering Types of problems}
  & Single objective   & 10,838 & Only one objective function is defined \\
  & Multi-objective    & 159   & Contains multiple objective functions \\
  & Multi-level        & 399   & Contains two or more levels of optimization \\
  & Uncertainty        & 158    & Contains some form of parameter or variable uncertainty \\
\midrule
\multirow{2}{*}{\centering Constraint availability}
  & Unconstrained & 155   & Constraints are absent \\
  & Constrained   & 11,399 & One or more constraints are present \\
\midrule
\multirow{2}{*}{\centering Model form}
  & General form  & 7,349 & Contains undefined parameters, functions, etc. \\
  & Fully defined & 4,205 & Completely defined with all necessary parameters \\
\midrule
\multirow{3}{*}{\centering Presentation form}
  & Vector form   & 608  & Defined in a form containing vector and matrix operations \\
  & Scaler form   & 10,246 & Defined in a form containing only scalar operations \\
  & Scalable form & 804  & Problem is scalable in terms of variables, objectives, etc. \\
\midrule
\multirow{8}{*}{\centering Other}
  & Linear             & 2,130 & All objectives and constraints are linear \\
  & Non-linear         & 9,122 & One or more objectives or constraints are non-linear \\
  & Continuous         & 10,806 & All variables and functions are continuous \\
  & Discontinuous      & 424  & Involves integer variables or contains discontinuities \\
  & Convex             & 2,580 & Belongs to the class of convex optimization \\
  & Non-convex         & 3,574 & Belongs to the class of non-convex optimization \\
  & Differentiable     & 9,502 & All the functions defined are differentiable \\
  & Non-differentiable & 502  & Some functions defined are not differentiable \\
\bottomrule
\multicolumn{4}{l}{\footnotesize There are formulations that belong to multiple categories and also formulations that cannot be classified appropriately.}
\end{tabular}
}
\end{table}

\begin{table}[h]
\centering
\caption{Sources used in creating \textit{AutoOpt-11k} dataset.}
\begin{tabular}{|p{4cm}|p{8.8cm}|}
\hline
\textbf{Source Type} & \textbf{References} \\
\hline
Books & \cite{floudas1990collection, bertsimas1997introduction,nocedal1999numerical,fletcher2000practical,rao2009engineering, guenin2014gentle, robinson2013introduction, fehr2007optimization, boyd2004convex, madsen2004optimization, floudas2013handbook, birge2011introduction, spall2003introduction, rao2009engineering, bazaraa2006nonlinear} \\
\hline
Research Papers and Reports & \cite{himmelblau1986nonlinear, bixby2007zib, dhyani2021alternate, ramamoorthy2018multiple, sinha2014test, huband2006review, andrei2008unconstrained, zitzler2000comparison, grosso2009finding, beasley1990or, caprara2000algorithms, israeli2002shortest, smith2008survey, ahmed2002sample, ben2009robust, pinedo1999operations, blazewicz1983scheduling, lukvsan2000test, Liang2006ProblemDA, stephanopoulos1975use, deb2006multi} \\
\hline
Mathematical Modeling Software Documentations & AMPL \cite{ampl2023manual}, GAMS \cite{gams2023manual}, PYOMO \cite{pyomo2023manual}, JuMP \cite{jump2023manual}, LINDO \cite{lindo2023manual} \\
\hline
Solver Documentations & CPLEX \cite{cplex-manual}, Gurobi \cite{gurobi-manual}, CBC \cite{cbc2023manual}, CLP \cite{forrest2005clp}, Ipopt \cite{wachter2006implementation} \\
\hline
Online Repositories & COIN-OR \cite{coinor2023website}, MIPLIB \cite{miplib2023}, OR-Library \cite{orlibrary2023}, UCI \cite{uci2023}, NEOS \cite{neos2023}, Netlib \cite{netlib2023} \\
\hline
\end{tabular}
\label{tab:autoopt_sources}
\end{table}

In the literature, mathematical models are expressed in a variety of notational and formatting styles. For instance, in the in-line style, the objective function and constraints are written horizontally in a single line, whereas in the multi-line style, they are arranged vertically with each component on a separate line. Additional stylistic variations include differences in objective function declarations (e.g., \textit{min/max} vs. \textit{minimize/maximize}), separators for constraints and objectives (e.g., \textit{s.t.}, \textit{w.r.t.}, \textit{subject to}), constraint indexing conventions (e.g., \( i \in [1, K] \) vs. \( i = 1, \dots, K \)), and text formatting aspects such as indentation, alignment, font type and size, line spacing, and use of bold or italic styles. Variations also occur in mathematical expressions (e.g., \( x^{1/2} \), \( x^{0.5} \), \( \sqrt{x} \)) and in variable or parameter naming conventions (e.g., \( p/q/r \), \( a/b/c \), \( x_1/x_2/x_3 \), or a mix such as \( p/a/x_1 \)). \textit{AutoOpt-11k} incorporates mathematical programs reflecting this broad spectrum of notational and stylistic differences. 

In the case of handwritten images, human involvement introduces additional layers of variability beyond those found in typeset representations. These include differences in handwriting style (e.g., font size, font type), paper type (such as plain or ruled), ink color (typically blue or black, and other colors on digital writing devices), and conditions under which the images are captured. Image captures vary in terms of camera angle, distance, orientation, lighting conditions, and camera specifications (such as resolution). Some of the images are also captured through a scanner. The dataset incorporates handwritten images reflecting this full range of variations that we obtained working with multiple annotators. Overall, 25 annotators, having engineering or business background and mathematics exposure up to the bachelors, were employed to form the dataset. We followed a two-phase process for dataset generation. In the first phase, 20 annotators identified the optimization problems from various sources and also submitted handwritten versions of some of those optimization problems. Thereafter, 5 annotators with background in programming, were recruited to prepare the LaTeX code of all the images and PYOMO script for a subset of images.

To minimize errors in dataset generation and ensure high-quality annotations, 
the annotators regenerated each image from the respective LaTeX code and visually compared the generated image against the original image. This cross-verification step ensured consistency between the image and its LaTeX representation, helping to identify and correct any discrepancies introduced during the initial annotation. 
In this second phase of annotation process, each of the 5 annotators (A1, A2, A3, A4, A5) annotated 30\% of the images and there was a 16.6\% overlap between any pair of annotators on average. The Inter Annotator Agreement (IAA) score for each pair of annotator is provided in Table \ref{tab:inter_annotator}. The reason for discrepancies between annotators was often because the code generated by them had syntactic differences for the same image.

\vspace{-2mm}
\begin{table*}[htbp]
\centering
\caption{Inter-Annotator Agreement Scores (BLEU and CER)}
\label{tab:inter_annotator}
\small 
\setlength{\tabcolsep}{3pt} 
\renewcommand{\arraystretch}{0.95} 
\begin{minipage}{0.48\textwidth}
  \centering
  \begin{tabular}{@{}crrrr@{}}
    \toprule
    \multirow{2}{*}{\textbf{Pair}} & \multicolumn{2}{c}{\textbf{BLEU}} & \multicolumn{2}{c}{\textbf{CER}} \\
    \cmidrule(lr){2-3} \cmidrule(lr){4-5}
     & \textbf{Mean} & \textbf{Std} & \textbf{Mean} & \textbf{Std} \\
    \midrule
    A1 vs A2 & 0.8187 & 0.1066 & 0.1784 & 0.1154 \\
    A1 vs A3 & 0.8185 & 0.1065 & 0.1788 & 0.1153 \\
    A1 vs A4 & 0.8195 & 0.1071 & 0.1776 & 0.1167 \\
    A1 vs A5 & 0.8201 & 0.1068 & 0.1782 & 0.1155 \\
    A2 vs A3 & 0.8588 & 0.1031 & 0.1267 & 0.1020 \\
    \bottomrule
  \end{tabular}
\end{minipage}
\hfill
\begin{minipage}{0.48\textwidth}
  \centering
  \begin{tabular}{@{}crrrr@{}}
    \toprule
    \multirow{2}{*}{\textbf{Pair}} & \multicolumn{2}{c}{\textbf{BLEU}} & \multicolumn{2}{c}{\textbf{CER}} \\
    \cmidrule(lr){2-3} \cmidrule(lr){4-5}
     & \textbf{Mean} & \textbf{Std} & \textbf{Mean} & \textbf{Std} \\
    \midrule
    A2 vs A4 & 0.8581 & 0.1042 & 0.1273 & 0.1040 \\
    A2 vs A5 & 0.8189 & 0.1062 & 0.1791 & 0.1148 \\
    A3 vs A4 & 0.8574 & 0.1044 & 0.1286 & 0.1050 \\
    A3 vs A5 & 0.8192 & 0.1064 & 0.1787 & 0.1150 \\
    A4 vs A5 & 0.8197 & 0.1070 & 0.1779 & 0.1159 \\
    \bottomrule
  \end{tabular}
\end{minipage}
\end{table*}

{\it AutoOpt-11k} dataset contains the LaTeX representation for all 11,554 images. The number of unique mathematical programs in the dataset is 7,637 out of 11,554, as we chose to include the typeset as well as handwritten versions of a variety of mathematical programs in our dataset. For a subset of 1,018 unique mathematical programs in the {\it AutoOpt-11k} dataset, we also provide the PYOMO scripts. Table~\ref{tab:dataset_stats} summarizes key statistics, and further details are available in Appendix~\ref{sec:Appendix_dataset}.

\vspace{-2mm}
\begin{table}[h!]
\centering
\caption{Summary Statistics for Image, LaTeX, and PYOMO Samples}
\scalebox{1}{
\begin{tabular}{lccccc}
\toprule
\textbf{Metric} & \textbf{Min} & \textbf{Max} & \textbf{Mean} & \textbf{Median} & \textbf{Total Samples} \\
\midrule
Image Width (px) & 159 & 3611 & 783.91 & 753.50 & \multirow{4}{*}{11,554} \\
Image Height (px) & 24 & 2670 & 338.89 & 295.00 & \\
Aspect Ratio (W/H) & 0.25 & 18.29 & 2.73 & 2.40 & \\
File Size (KB) & 3.06 & 1399.98 & 95.58 & 41.68 & \\
\midrule
LaTeX Length (chars) & 14 & 1,620 & 212.23 & 180.00 & 11,554 \\
PYOMO Length (chars) & 192 & 1,087 & 390.30 & 362.00 & 1,018 \\
\bottomrule
\end{tabular}
}
\label{tab:dataset_stats}
\end{table}

\vspace{-4mm}
\section{Framework for Automating Optimization Problem Solving}\label{sec:AutoOpt}
This section details the development of \textit{AutoOpt} framework, illustrated in Figure~\ref{fig:AutoOpt}. The framework is composed of three sequential modules—M1, M2, and M3—each responsible for a specific task in the automated optimization pipeline. The output from each module serves as the input for the subsequent one. 
Details on computational set up and infrastructure, along with detailed results from multiple runs are relegated to Appendices~\ref{sec:Append_m1}, \ref{sec:Append_m2} and \ref{sec:Append_m3}.

\subsection{Module M1: Image to LaTeX Code Generation}\label{sec:Module_M1}
\begin{figure*}[hbt]
\centering
\includegraphics[width=1.02\textwidth]{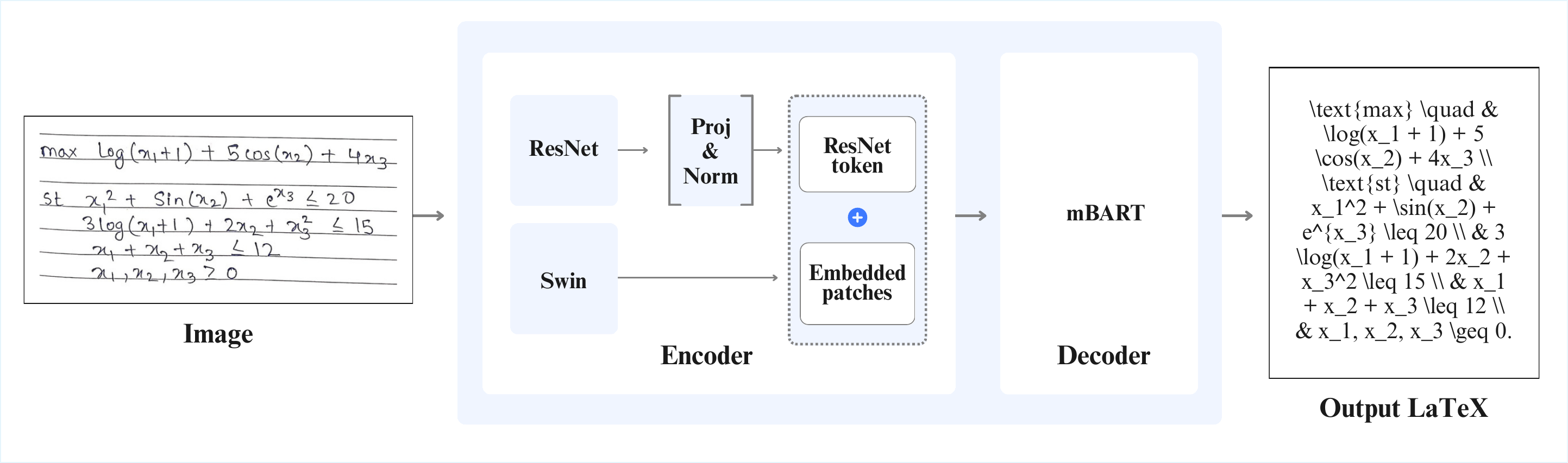}
\caption{Architecture of the deep learning model developed for MER}
\label{fig:module_m1}
\end{figure*}

In this module, we propose a hybrid deep learning architecture suitable for MER. The proposed model extends the NOUGAT architecture~\cite{blecher2023nougat, xu2025nougatLaTeXocr}, a DONUT-based framework comprising a vision encoder and a text decoder, specifically designed for typeset scientific documents. Given the inherent complexity of two-dimensional mathematical notation, which may be handwritten or typeset, we design a hybrid vision encoder by integrating ResNet and Swin Transformer~\cite{liu2021swin} thereby leveraging the strengths of both Convolutional Neural Networks (CNNs) and Transformers~\cite{peng2021conformer}. CNNs are effective at capturing local visual patterns, while Transformers excel at modeling long-range dependencies and global structure. 
The overall architecture is shown in Figure~\ref{fig:module_m1}.

The hybrid encoder combines ResNet-101 and Swin Transformer. ResNet-101 serves as a backbone for local feature extraction, producing a 2048-dimensional feature vector via average pooling. These features capture fine-grained local characteristics such as symbol shape and stroke patterns—crucial for both printed and handwritten expressions. 
In parallel, the Swin Transformer processes the input image by applying hierarchical self-attention within and across non-overlapping windows. This enables it to capture long-range dependencies and spatial layouts such as superscripts, subscripts, matrices, and fractions. The output of the Swin Transformer is a sequence of patch-level embeddings. 
To combine both streams, the ResNet-generated feature vector is projected to match the Transformer’s hidden dimension and prepended to the sequence of patch embeddings, enabling joint learning of global and local context. To integrate CNN-derived features with Transformer-based patch embeddings, we introduce a lightweight fusion strategy provided below. 
\begin{align*}
\mathbf{f}_{\text{ResNet}} = \alpha \cdot \text{LayerNorm}(\text{Proj}(\text{ResNet}_{\text{feat}})),
\end{align*}
where $\alpha \in \mathbb{R}$ is a learnable scalar initialized to zero, acting as a gating parameter during early training. This vector is then prepended to the sequence of Swin Transformer embeddings.

The decoder in our proposed architecture is based on the mBART~\cite{liu2020multilingual,lewis2019bart} architecture—a pre-trained Transformer-based autoregressive decoder. The decoder generates LaTeX code token-by-token, attending to the fused encoder outputs via cross-attention and to past generated tokens via causal self-attention. The NOUGAT model uses the same decoder, therefore we initialize the decoder with pre-trained NOUGAT weights while training it on our task-specific dataset.

\subsubsection{Experimental Results}\label{sec:expt_m1}
To ensure uniformity in input dimensions and improve model robustness, we implement a tailored preprocessing pipeline. Each input image is first resized so that its longer side fits within a $768 \times 1024$ canvas while maintaining the aspect ratio. The resized image is then center-padded on a white background to match the target dimensions. This standardization ensures consistent input representation regardless of the original aspect ratio. Given the complexity and density of the mathematical expressions in our dataset, we apply contrast enhancement to improve visual clarity. Additionally, an unsharp mask filter is used to accentuate symbol boundaries and fine pen strokes, which are critical for handwritten mathematical notation.

We adopt a transfer learning approach to train our model.
ResNet-101 is initialized with ImageNet weights~\cite{deng2009imagenet}, while the Swin Transformer and mBART decoder are initialized using pre-trained weights from the NOUGAT model. These pre-trained models, trained on large-scale scientific corpora, provide a good starting point, enabling faster convergence and better generalization while training on {\it AutoOpt-11k}. In our experiments a training, validation, and test split of 80\%, 10\%, and 10\% is used.

\begin{figure}[htbp]
    \centering
    \includegraphics[width=0.48\textwidth]{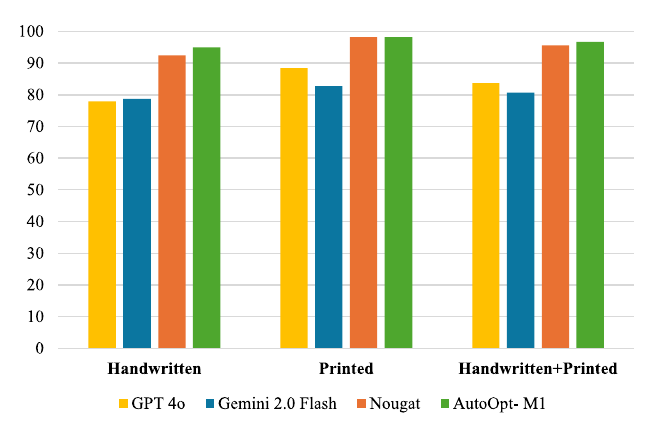}
    \caption{BLUE Score}
    \label{fig:Blue_Score}
    \vspace{-6mm}
\end{figure}

We compare our model that we refer to as AutoOpt-M1 against Nougat, ChatGPT and Gemini. Nougat has been fine-tuned on {\it AutoOpt-11k} dataset. For ChatGPT we use the GPT 4o model, and for Gemini we use Gemini 2.0 Flash model, both through their APIs. 
The ChatGPT and Gemini models were not fine-tuned but were given appropriate prompts with examples. Figure~\ref{fig:Blue_Score} compares these models with respect to BLUE Score (larger is better), and Table~\ref{tab:metrics} compares them with respect to Character Error Rate (smaller is better).
Clearly, Nougat and AutoOpt-M1 are much smaller and better performing models as compared to ChatGPT and Gemini models. Between Nougat and AutoOpt-M1, the latter outperforms the prior on all metrics except on Character Error Rate for Printed. However, note that there is a possibility to produce slightly different LaTeX code for the same image; therefore, for all models there are situations where the predicted LaTeX code is semantically correct, but the ground truth LaTeX is different. For further details, refer to Appendix~\ref{sec:Append_m1}.

\vspace{-2mm}
\begin{table}[htbp]
\centering
\caption{Performances of various approaches considered for MER task}\label{tab:metrics}
\vspace{1mm}
\begin{tabular}{|c|p{1.4cm}|p{1.4cm}|p{1.5cm}|}
\hline
\textbf{Model} 
& \textbf{HW} 
& \textbf{PR} 
& \textbf{HW+PR} \\
\cline{2-4}
\textbf{(Model Size)} & \multicolumn{3}{|c|}{\textbf{Character Error Rate}} \\
\hline
GPT 4o (Large) & 0.1465 & 0.0664 & 0.1017 \\
Gemini 2.0 Flash (Large) & 0.1607 & 0.1047 & 0.1338 \\
Nougat (348.7M) & 0.0752 & \textbf{0.0168} & 0.0440 \\
AutoOpt-M1 (393.3M) & \textbf{0.0412} & 0.0176 & \textbf{0.0286} \\
\hline
\multicolumn{4}{|l|}{\fontsize{8.5}{8}\selectfont HW: Handwritten; PR: Printed; HW+PR: Handwritten+Printed}\\ 
\cline{1-4}
\end{tabular}
\end{table}

We arrived at the hybrid encoder architecture based on an ablation study, in which we tried Deep Learning (DL) models with different architectures such as DL1 (with CNN, without Transformer): BLEU- 16.10, CER- 0.8812; DL2 (without CNN, with Transformer): BLEU- 95.51, CER- 0.0440; and DL3 (with CNN, with Transformer): BLEU- 96.70, CER- 0.0286. Finally, based on the comparison of BLEU and CER performance metrics, DL3 (Figure \ref{fig:module_m1}) is selected.

\subsection{Module M2: LaTeX to PYOMO Script Generation}\label{sec:Module_M2}
This module generates the model-specific PYOMO script from LaTeX code. To operationalize this task, we fine-tune a causal decoder-only transformer model using the instruction-style data. We specifically considered the DeepSeek-Coder 1.3B \cite{guo2024deepseek}, a pre-trained language model as the base. This instruct model has strong code generation capability and pre-trained alignment to instruction-following tasks, and it is smaller as compared to other coding LLMs. Fine-tuning is performed on 80\% of 1,018 mathematical models, while the remaining 20\% is used for testing. We refer to the fine-tuned model as AutoOpt-M2, for which we obtained a BLUE Score of 88.25 and Character Error Rate of 0.0825. Refer to Appendix~\ref{sec:Append_m2} for additional details.


\subsection{Module M3: Optimization using Bilevel Optimization based Decomposition Method}\label{sec:Module_M3}
In module M3, we implement an optimization method capable of efficiently solving a wide range of small-to-large scale optimization problems. Based on the nature of delivered solution (i.e., optimal or approximate), optimization methods are broadly classified into two categories: exact and approximation methods. Classical mathematical programming based techniques (such as linear programming \cite{dantzig2016linear}, integer programming \cite{wolsey2020integer}, etc.) fall into the category of exact methods. These methods guarantee optimality but often require certain regularities, like linearity, continuity, differentiability, etc. On the other side, approximate methods like heuristics and metaheuristics can lead to a satisfactory solution on irregular problems but may not scale well and do not guarantee optimality. Interestingly, some recent studies \cite{liu2024EoH,wu2024evolutionary,yao2025multi_AAAI} explore how LLMs can be used to design problem-specific heuristics. An alternative line of study \cite{sinha2024decomposition} proposes a bilevel optimization based decomposition strategy to utilize metaheuristic and classical approaches simultaneously to solve a wide variety of problems. Our implementation in module M3 is an extension of work by Sinha et al. \cite{sinha2024decomposition}.

\begin{figure*}[hbt]
\centering
\includegraphics[width=0.7\textwidth]{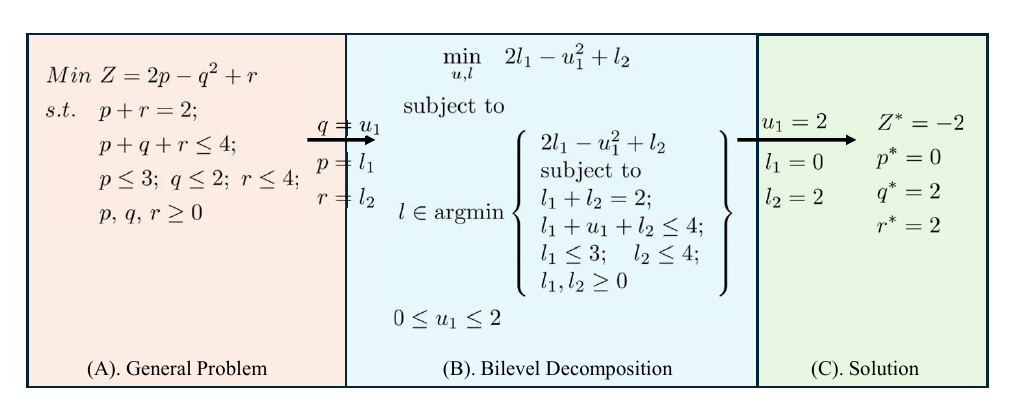}
\caption{Problem solving using Bilevel optimization-based decomposition approach}
\vspace{-2mm}
\label{fig:Inst_solve}
\end{figure*}

In Figure \ref{fig:Inst_solve}, we demonstrate the procedure of BOBD method by solving the optimization problem considered in Figure \ref{fig:AutoOpt}. The optimization problem in the first tab of Figure~\ref{fig:Inst_solve} contains three variables $p,q$ and $r$. This is a non-convex optimization problem\footnote{Note that despite the example being a non-convex problem it can be solved efficiently by exact methods because of the quadratic nature of the objective function and linear nature of the constraints. However, we have chosen this problem for the ease of discussion.} because of the presence of the term $-q^2$ in the objective function that is to be minimized. However, note that if the value if $q$ is fixed, this problem becomes a linear program. By using an intelligent sampling approach like a metaheuristic for $q$ and solving a linear program with respect to $p$ and $r$ for each sample of $q$ one can find an approximate optimal solution. Such a decomposition approach breaks the problem into a bilevel optimization structure where the upper level is handled by one optimization algorithm and the lower (nested) level is handled by another optimization algorithm. The second tab in Figure~\ref{fig:Inst_solve} shows how the same problem can be written as a bilevel optimization problem by representing $q$ as $u_1$, $p$ as $l_1$ and $r$ as $l_2$. In our implementation of BOBD, we use a genetic algorithm at the upper level and rely on a convex optimization solver at the lower level. Within the iterations of the genetic algorithm, we classify the variables into upper and lower levels using a machine learning approach. 

\vspace{-2mm}

Note that in module M3, one can use any other approach for solving the optimization problem specified in the PYOMO script. However, we use the BOBD approach because it allows one to capitalize on two mathematical optimization paradigms simultaneously to solve optimization problems, thereby reducing human intervention. Additional implementation details of the BOBD approach and results are provided in Appendix~\ref{sec:Append_m3}. 
The results demonstrate superior performance of BOBD approach in handling a large variety of problems compared to other popular approaches. However, note that optimization methods are evaluated on optimality guarantees and convergence rates, and we do not make any claims of our BOBD implementation being better than any specialized optimization technique on these aspects.

\subsection{Performance of AutoOpt Framework}\label{sec:AutoOpt_performance}
The performance of \textit{AutoOpt} framework is evaluated using two approaches: (i) module-level evaluation and (ii) framework-level evaluation. In module-level evaluation, we consider the individual performance of each module to estimate the reliability of entire framework. For module M1, the Character Error Rate (CER) is 0.0286; hence, reliability of module M1 is estimated as (1-CER)$\times$100 = (1-0.0286)$\times$100 = 97.14\%. For module M2, CER rate is 0.0825; hence, reliability of module M2 is estimated as (1-0.0825)$\times$100 = 91.75\%. Module M3 contains the optimization solver that solves exactly what is provided in PYOMO script, i.e., there is no prediction task or error-prone task associated with this module. Thus, the reliability or success-rate of entire \textit{AutoOpt} framework can be estimated as (0.9714$\times$0.9175)$\times$100= 89.12\%. This estimate is actually a lower bound, as in many cases where the LaTeX or PYOMO is syntactically different from the expected output, the CER metric incorrectly counts such differences as character errors. 

In framework-level evaluation, we measure the performance of the complete pipeline (M1–M2–M3) on 500 sample problems outside the {\it AutoOpt-11k} dataset. The overall success rate (i.e., ability to correctly read the problem in LaTeX and PYOMO and subsequently deploy the solver successfully) was observed to be 94.20\%.    

\section{Conclusions}\label{sec:conclusions}
This study introduces {\it AutoOpt}, an end-to-end automated framework that enables optimization problem-solving directly from images of mathematical formulations, thereby significantly reducing human intervention. Central to this framework is {\it AutoOpt-11k}, a curated dataset comprising over 11,554 images of handwritten and typeset mathematical programs, labeled with corresponding LaTeX code for all images and modeling language script for a subset of images. This dataset addresses a longstanding gap in image-based optimization data resources that has the potential to automate optimization problem solving.

The proposed framework consists of three integrated modules: M1 (Image-to-LaTeX), M2 (LaTeX-to-PYOMO), and M3 (Optimization Solver). Each module is powered by custom-developed or fine-tuned deep learning and optimization methods, achieving strong performance across tasks. The deep learning model in M1 outperforms the existing state-of-the-art tools like ChatGPT, Gemini, and Nougat. Additionally, the BOBD method in M3 demonstrates superior performance in solving a wider variety of optimization problems compared to other approaches. By automating the complete pipeline—from image acquisition to solution generation—{\it AutoOpt} framework offers a powerful and accessible solution for both researchers and practitioners. The public release of the dataset and the framework is expected to encourage future research at the intersection of computer vision, natural language processing, and mathematical optimization. Future research will also address some of the limitations of this study, for instance, handling ill-defined optimization problems effectively, or handling optimization problem definitions that span multiple pages or images.


\bibliographystyle{plain}


\newpage
\appendix
\newpage

\section{Appendix: Dataset}\label{sec:Appendix_dataset}
\begin{figure}[h]
        \centering
        \includegraphics[width=\linewidth]{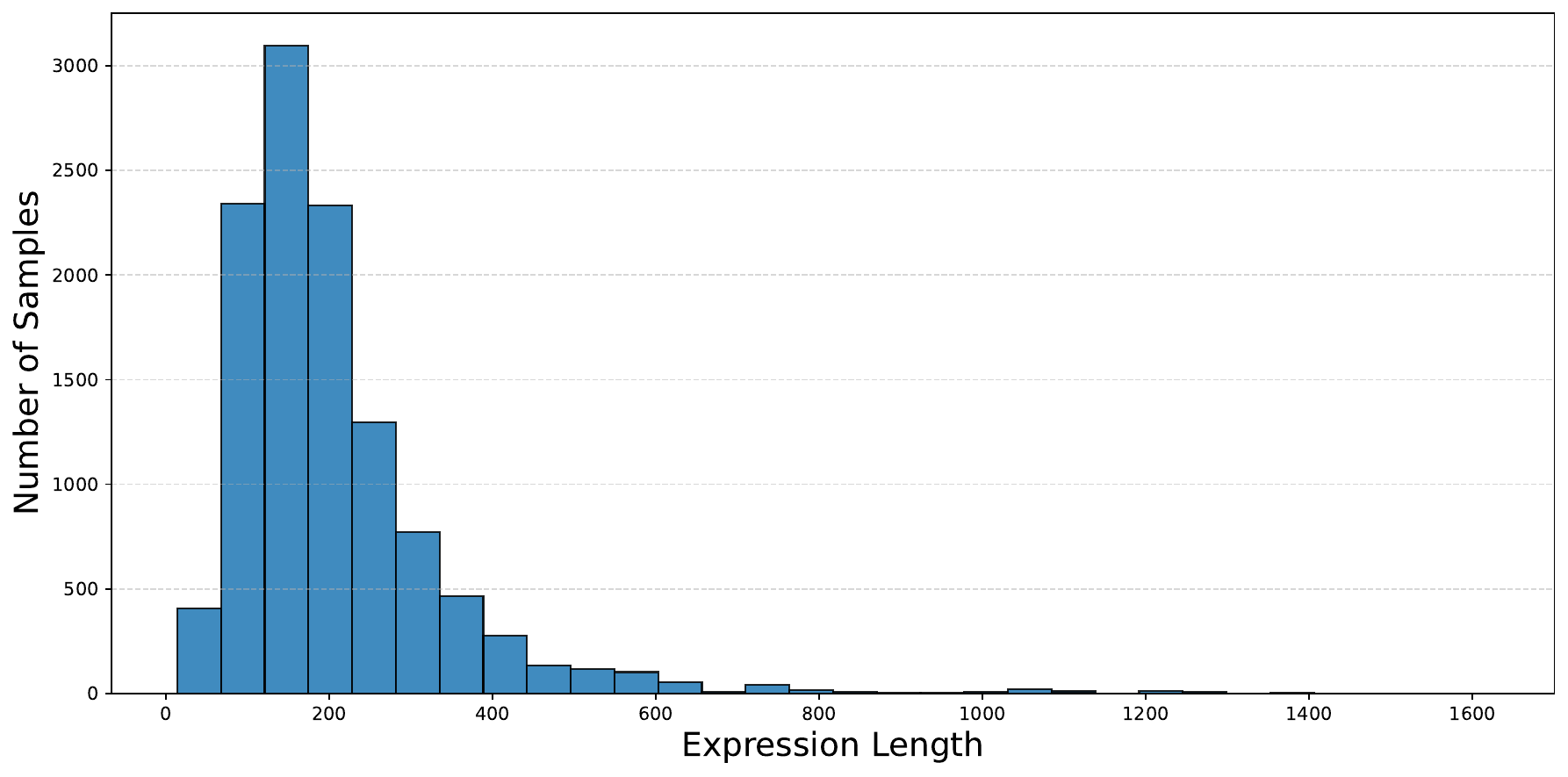}
        \caption{Histogram of LaTeX expression lengths}
        \label{fig:dataset_figure1}
    \end{figure}

Figures~\ref{fig:dataset_figure1}, \ref{fig:dataset_figure2}, and~\ref{fig:dataset_figure3} offer deeper insights into the structure and content of the \textit{AutoOpt-11k} dataset developed in this study. These visualizations collectively help characterize the dataset along multiple dimensions—expression complexity, comparative scale, and token diversity.

Figure~\ref{fig:dataset_figure1} presents a histogram of LaTeX expression lengths, indicating the number of samples corresponding to different expression sizes. This figure is particularly useful for understanding the distribution of expression complexity in our dataset. Unlike many datasets that contain shorter and simpler expressions, our dataset encompasses a broad range of lengths, including a substantial number of longer and more elaborate expressions. This distribution is indicative of real-world mathematical programs.
Figure~\ref{fig:dataset_figure2} provides a comparative analysis of LaTeX expression lengths across multiple publicly available datasets \cite{aida_calculus_dataset,xie2023icdar,deng2017image,gervais2024mathwriting} for mathematical expression recognition. It is evident from the figure that our dataset distinguishes itself by including a significantly higher proportion of longer, multi-line expressions. This unique characteristic enhances its applicability to practical use cases that require parsing long mathematical expressions. 
Figure~\ref{fig:dataset_figure3} illustrates the 100 most frequent LaTeX tokens in our dataset, underscoring its syntactic richness and diversity. The tokens span a wide range of categories, including comparison operators, set theory notations, mathematical operators, syntactic elements, Latin letters, numbers, Blackboard capital letters, Greek symbols, mathematical constructs, modifiers, matrix environments, delimiters, arrows, dots, punctuations and various other symbols. This token diversity confirms the dataset’s relevance for training models that must generalize across diverse types of notation.

Furthermore, Table ~\ref{tab:dataset_all1} and Table ~\ref{tab:dataset_all2} provide concrete examples from the dataset, showcasing images of optimization formulations along with their corresponding LaTeX and PYOMO representations. These examples demonstrate the alignment between visual representations and their semantic counterparts, highlighting the dataset's utility for a variety of tasks. 
Together, the figures and tables substantiate the comprehensiveness and quality of our dataset, validating its potential to support robust training and evaluation of machine learning models targeting advanced mathematical understanding and code generation tasks.

\vspace{1cm}

\newpage
\begin{landscape}
\begin{figure}[htbp]
    \centering

    \begin{minipage}[t]{0.68\linewidth}
        \centering
        \includegraphics[width=\linewidth]{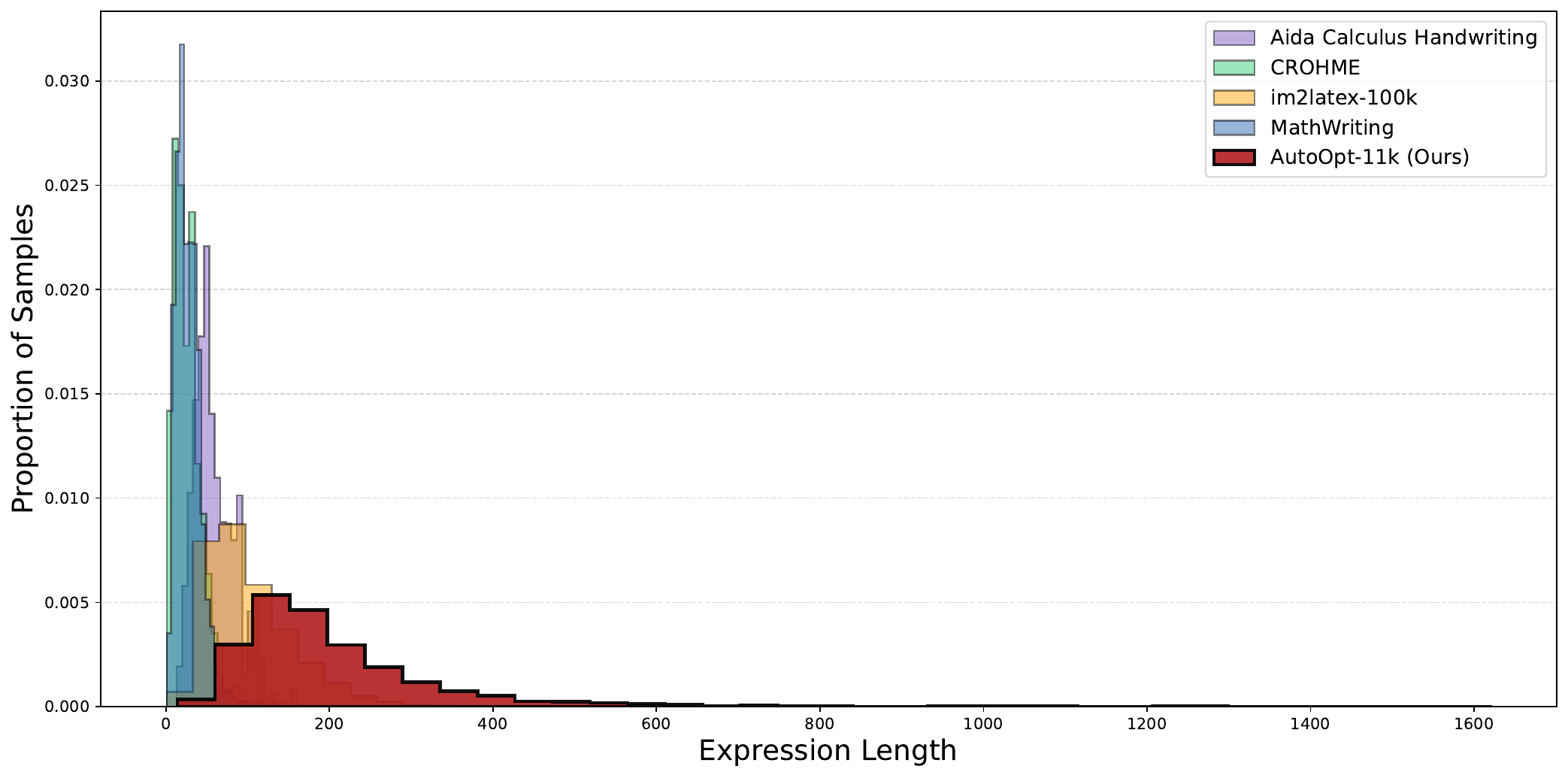}
        \caption{Normalized comparison of LaTeX expression lengths}
        \label{fig:dataset_figure2}
    \end{minipage}

    \vspace{6mm} 

    \begin{minipage}[t]{0.98\linewidth}
        \centering
        \includegraphics[width=\linewidth]{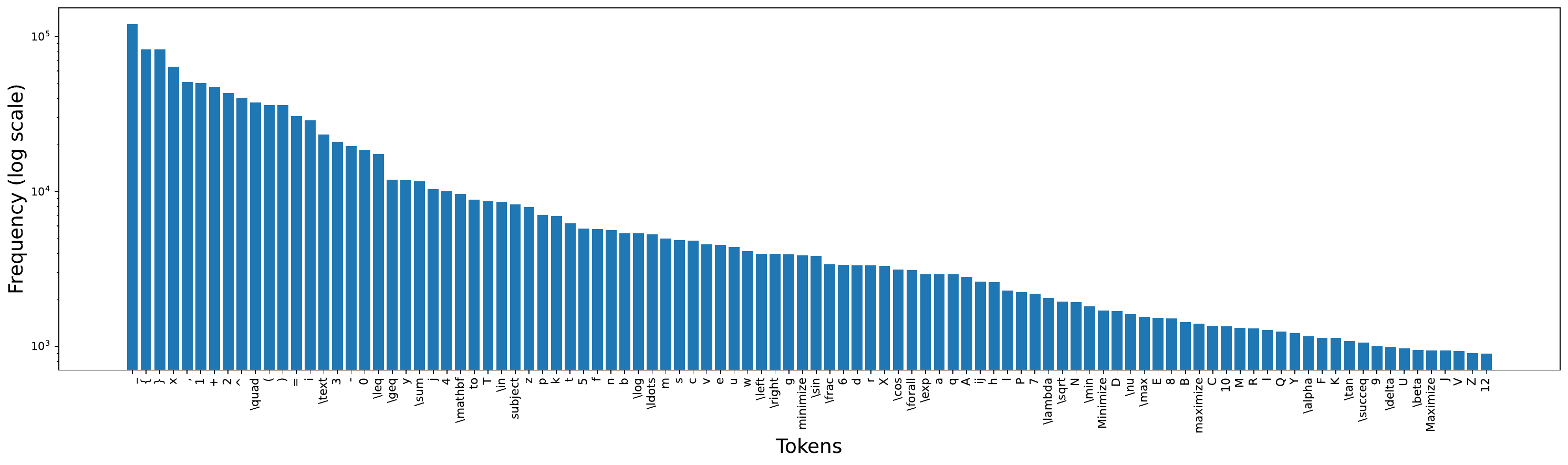}
        \caption{Top 100 most frequent tokens in LaTeX}
        \label{fig:dataset_figure3}
    \end{minipage}
\end{figure}

\begin{table}[h]
\scriptsize
\caption{Samples from \textit{AutoOpt-11k} dataset: Images and Labels}
\label{tab:dataset_all1}
\begin{flushleft}
\begin{tabular}{|c|p{17cm}|}
\hline
\textbf{\rule{0pt}{1.2em}Image} & \textbf{\rule{0pt}{1.2em}LaTex and PYOMO}
\\ \hline \hline
\multirow{3}{*}{\raisebox{-1.2\height}{\includegraphics[width=5cm]{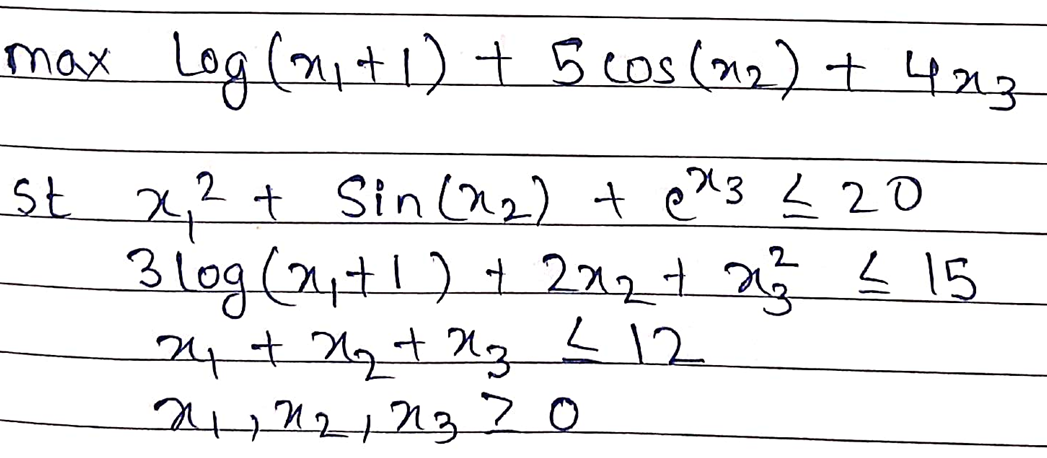}}} 
& \vtop{%

\begin{verbatim}
\text{max} \quad & \log(x_1 + 1) + 5 \cos(x_2) + 4x_3 \\ \text{st} \quad & x_1^2 + \sin(x_2) + e^{x_3} \leq 20 \\ & 3 \log(x_1 + 1) 
+ 2x_2 + x_3^2 \leq 15 \\ & x_1 + x_2 + x_3 \leq 12 \\ & x_1, x_2, x_3 \geq 0.
\end{verbatim}

} \\
\cline{2-2}
& \vtop{%
\begin{verbatim}
model.x1 = Var(within=NonNegativeReals)\nmodel.x2 = Var(within=NonNegativeReals)\nmodel.x3 = Var(within=NonNegativeReals)\n
def objective_function(model):\n return log(x1 + 1) + 5*cos(x2) + 4*x3\nmodel.OF = Objective(rule=objective_function,sense=maximize)
\nmodel.Constraint1 = Constraint(expr = x1**2 + sin(x2) + exp(x3) - 20 <= 0)\nmodel.Constraint2 = Constraint(expr = 3*log(x1 + 1) + 
2*x2 + x3**2 - 15 <= 0)\nmodel.Constraint3 = Constraint(expr = x1 + x2 + x3 - 12 <= 0)
\end{verbatim}
} \\
\hline 
\multirow{3}{*}{\raisebox{-1.1\height}{\includegraphics[width=5cm]{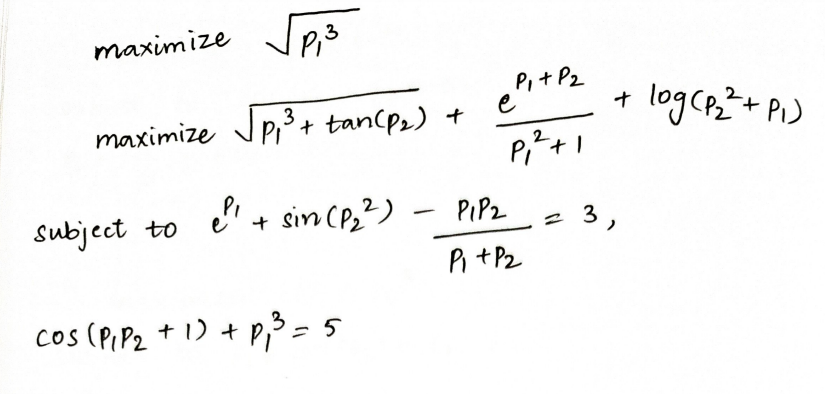}}} 
& \vtop{%

\begin{verbatim}
\text{maximize} \quad & \sqrt{p_1^3 + \tan(p_2)} + \frac{e^{p_1 + p_2}}{p_1^2 + 1} + \log(p_2^2 + p_1) \\ \text{subject to} \quad & 
e^{p_1} + \sin(p_2^2) - \frac{p_1 p_2}{p_1 + p_2} = 3, \\ & \cos(p_1 p_2 + 1) + p_1^3 = 5
\end{verbatim}

} \\
\cline{2-2}
& \vtop{%
\begin{verbatim}
model.p1 = Var()\nmodel.p2 = Var()\ndef objective_rule(model):\n return sqrt(p1**3 + tan(p2)) + exp(p1 + p2)/(p1**2 + 1) +
log(p2**2 + p1)\n model.objective = Objective(rule=objective_rule, sense=maximize)\nmodel. Constraint1 = Constraint(expr = exp(p1) 
+ sin(p2**2) - (p1*p2)/(p1 + p2) == 3)\n model.Constraint2 = Constraint(expr = cos(p1*p2 + 1) + p1**3 == 5)
\end{verbatim}
} \\ 
\hline 
\multirow{3}{*}{\raisebox{-1.2\height}{\includegraphics[width=5cm]{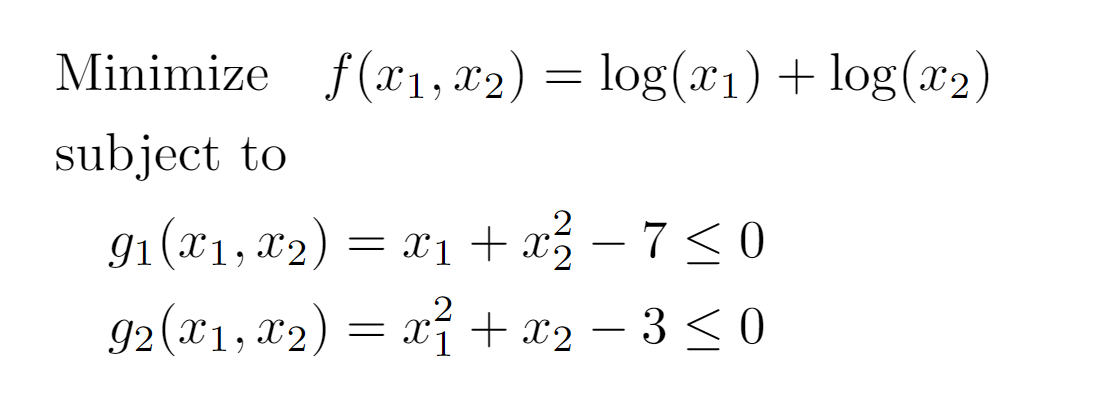}}} 
& \vtop{%

\begin{verbatim}
& \text{Minimize} \quad f(x_1, x_2) = \log(x_1) + \log(x_2) \\ & \text{subject to} \\ & \quad g_1(x_1, x_2) = x_1 + x_2^2 
- 7 \leq 0 \\ & \quad g_2(x_1, x_2) = x_1^2 + x_2 - 3 \leq 0
\end{verbatim}

} \\
\cline{2-2}
& \vtop{%
\begin{verbatim}
model.x1 = Var(within=PositiveReals)\nmodel.x2 = Var(within=PositiveReals)\ndef objective_function(model):\n return log(x1) + 
log(x2)\nmodel.obj = Objective(rule=objective_function, sense=minimize)\nmodel.Constraint1 = Constraint(expr = x1 + x2**2 <= 7)\n
model.Constraint2 = Constraint(expr = x1**2 + x2 <= 3)
\end{verbatim}
} \\ 
\hline 
\multirow{3}{*}{\raisebox{-1.1\height}{\includegraphics[width=5cm]{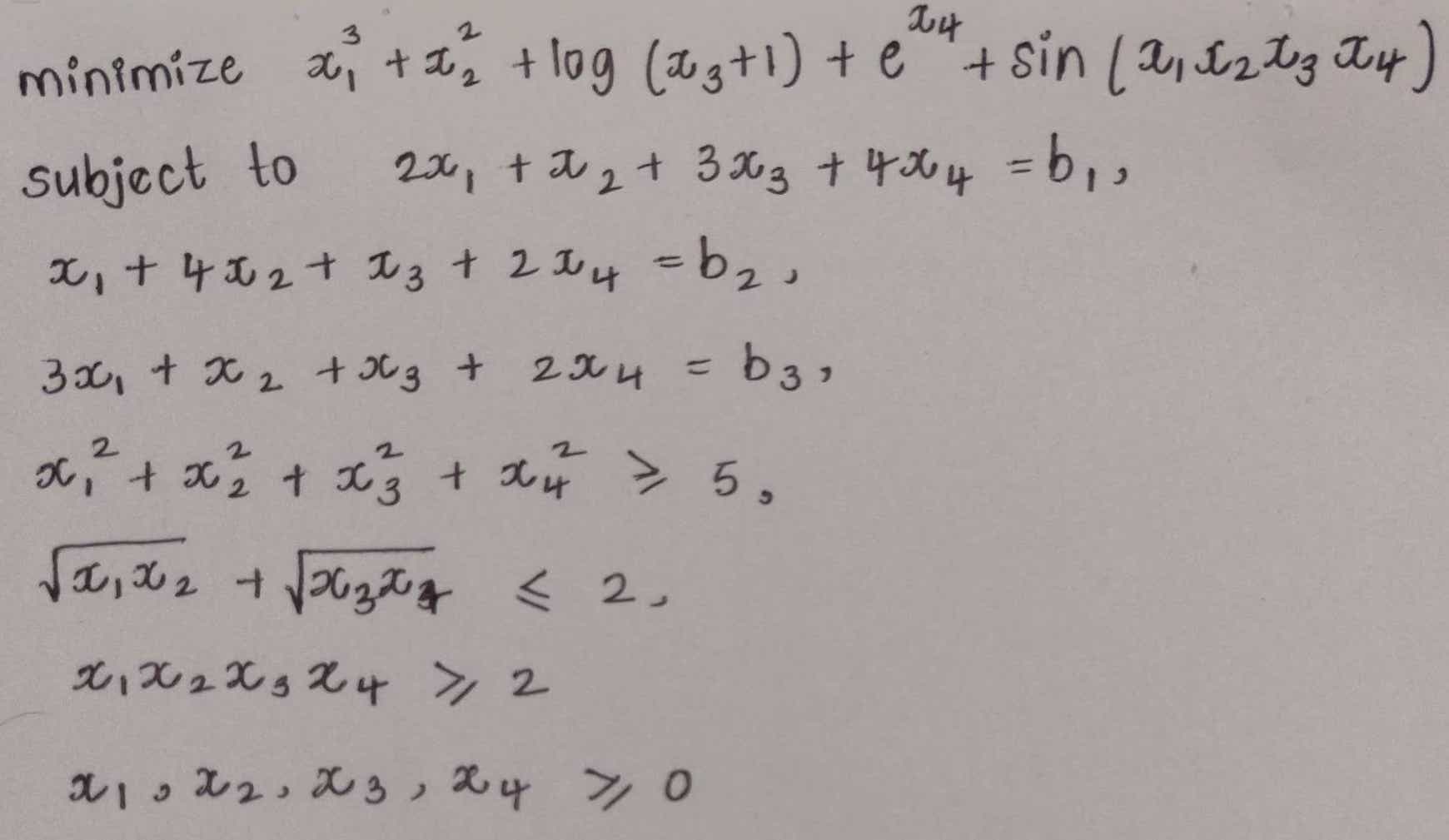}}} 
& \vtop{%

\begin{verbatim}
\text{minimize} \quad & x_1^3 + x_2^2 + \log(x_3 + 1) + e^{x_4} + \sin(x_1 x_2 x_3 x_4) \\ \text{subject to} \quad & 2x_1 + x_2 + 
3x_3 + 4x_4 = b_1, \\ & x_1 + 4x_2 + x_3 + 2x_4 = b_2, \\ & 3x_1 + x_2 + x_3 + 2x_4 = b_3, \\ & x_1^2 + x_2^2 + x_3^2 + x_4^2 \geq 
5, \\ & \sqrt{x_1 x_2} + \sqrt{x_3 x_4} \leq 2, \\ & x_1 x_2 x_3 x_4 \geq 2, \\ & x_1, x_2, x_3, x_4 \geq 0
\end{verbatim}

} \\
\cline{2-2}
& \vtop{%
\begin{verbatim}
model.x1 = Var(within=NonNegativeReals)\nmodel.x2 = Var(within=NonNegativeReals)\nmodel.x3 = Var(within=NonNegativeReals)\n
model.x4 = Var(within=NonNegativeReals)\ndef objective_rule(model):\n    return x1**3 + x2**2 + log(x3 + 1) + exp(x4) + 
sin(x1 * x2 * x3 * x4)\nmodel.objective = Objective(rule=objective_rule, sense=minimize)\nmodel.Constraint1 = Constraint(expr= 2*x1 
+ x2 + 3*x3 + 4*x4 == b1)\nmodel.Constraint2 = Constraint(expr = x1 + 4*x2 + x3 + 2*x4 == b2)\nmodel.Constraint3 = Constraint(expr= 
3*x1 + x2 + x3 + 2*x4 == b3)\nmodel.Constraint4 = Constraint(expr = x1**2 + x2**2 + x3**2 + x4**2 >= 5)\n model.Constraint5 = 
Constraint(expr= sqrt(x1 * x2) + sqrt(x3 * x4) <= 2)\nmodel.Constraint6 = Constraint(expr= x1 * x2 * x3 * x4 >= 2)
\end{verbatim}
} \\ 
\hline
\end{tabular}
\end{flushleft}
\end{table}

\begin{table}[h]
\scriptsize
\caption{Samples from \textit{AutoOpt-11k} dataset: Images and Labels (Cont.)}
\label{tab:dataset_all2}
\begin{flushleft}
\begin{tabular}{|c|p{17cm}|}
\hline
\textbf{\rule{0pt}{1.2em}Image} & \textbf{\rule{0pt}{1.2em}LaTex and PYOMO}
\\ \hline \hline
\multirow{3}{*}{\raisebox{-1.1\height}{\includegraphics[width=5cm]{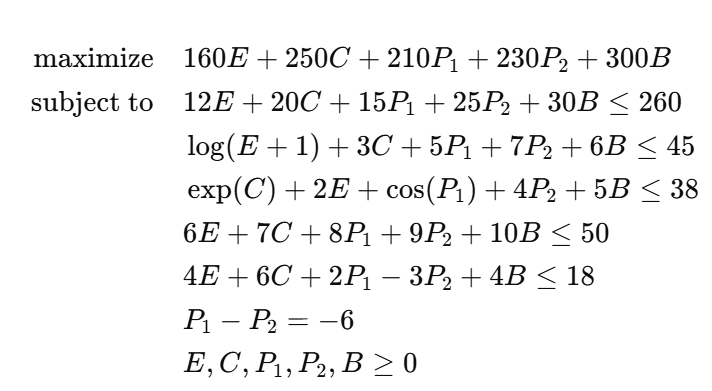}}} 
& \vtop{%

\begin{verbatim}
\text{maximize} & 160E + 250C + 210P_1 + 230P_2 + 300B \\ \text{subject to} & 12E + 20C + 15P_1 + 25P_2 + 30B \leq 260 \\ &
\log(E + 1) + 3C + 5P_1 + 7P_2 + 6B \leq 45 \\ & \exp(C) + 2E + \cos(P_1) + 4P_2 + 5B \leq 38 \\ & 6E + 7C + 8P_1 + 9P_2 + 10B \leq
50 \\ & 4E + 6C + 2P_1 - 3P_2 + 4B \leq 18 \\ & P_1 - P_2 = -6 \\ & E, C, P_1, P_2, B \geq 0.
\end{verbatim}

} \\
\cline{2-2}
& \vtop{%
\begin{verbatim}
model.E = Var(within=NonNegativeReals)\nmodel.C = Var(within=NonNegativeReals)\nmodel.P1 = Var(within=NonNegativeReals)\nmodel.P2 = 
Var(within=NonNegativeReals)\nmodel.B = Var(within=NonNegativeReals)\ndef objective_function(model):\n    return 160*E + 250*C + 
210*P1 + 230*P2 + 300*B\nmodel.obj = Objective(rule=objective_function, sense=maximize)\nmodel.Constraint1 = Constraint(expr = 12*E 
+ 20*C + 15*P1 + 25*P2 + 30*B <= 260)\nmodel.Constraint2 = Constraint(expr = log(E + 1) + 3*C + 5*P1 + 7*P2 + 6*B <= 45)\n
model.Constraint3 = Constraint(expr = exp(C) + 2*E + cos(P1) + 4*P2 + 5*B <= 38)\nmodel.Constraint4 = Constraint(expr = 6*E + 7*C + 
8*P1 + 9*P2 + 10*B <= 50)\nmodel.Constraint5 = Constraint(expr = 4*E + 6*C + 2*P1 - 3*P2 + 4*B <= 18)\nmodel.Constraint6 = 
Constraint(expr = P1 - P2 == -6)
\end{verbatim}
} \\
\hline 
\multirow{3}{*}{\raisebox{-1.3\height}{\includegraphics[width=5cm]{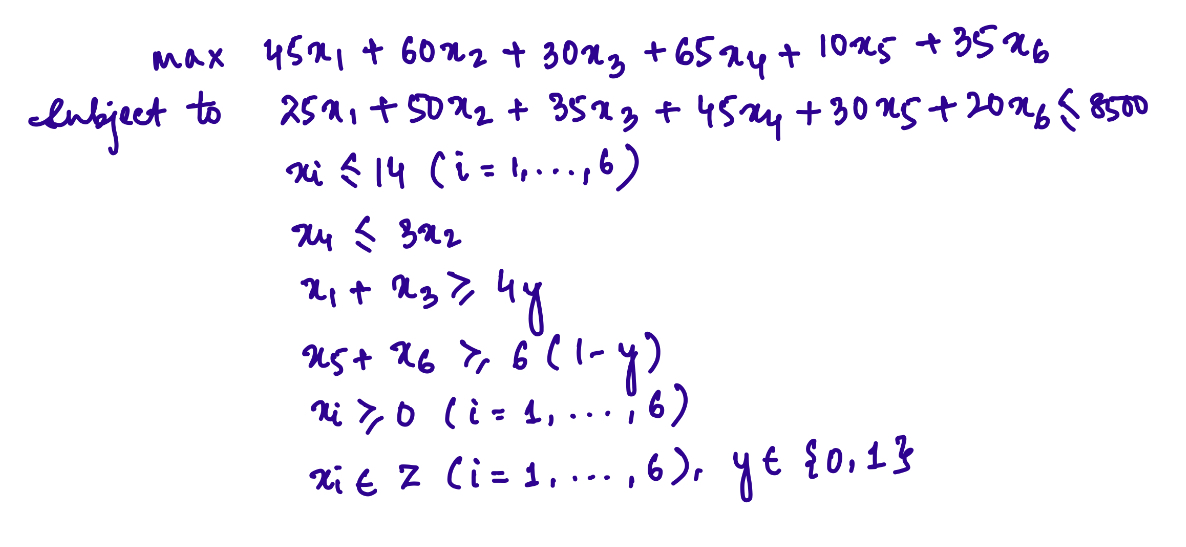}}} 
& \vtop{%

\begin{verbatim}
\text{max} \quad & 45x_1 + 60x_2 + 30x_3 + 65x_4 + 10x_5 + 35x_6 \\ \text{subject to} \quad & 25x_1 + 50x_2 + 35x_3 + 45x_4 + 30x_5 
+ 20x_6 \leq 8500 \\ & x_i \leq 14 \quad (i = 1, \ldots, 6) \\ & x_4 \leq 3x_2 \\ & x_1 + x_3 \geq 4y \\ & x_5 + x_6 \geq 6(1 - y)
\\ & x_i \geq 0 \quad (i = 1, \ldots, 6) \\ & x_i \quad \text{integer} \quad (i = 1, \ldots, 6) \\ & y \in \{0, 1\}.
\end{verbatim}

} \\
\cline{2-2}
& \vtop{%
\begin{verbatim}
model.x1 = Var(within=NonNegativeReals, domain=Integers)\\nmodel.x2 = Var(within=NonNegativeReals, domain=Integers)\\nmodel.x3 = 
Var(within=NonNegativeReals, domain=Integers)\\nmodel.x4 = Var(within=NonNegativeReals, domain=Integers)\\nmodel.x5 = Var(within=
NonNegativeReals, domain=Integers)\\nmodel.x6 = Var(within=NonNegativeReals, domain=Integers)\\nmodel.y = Var(domain=Binary)\\n
def objective_function(model):\\n    return 45*x1 + 60*x2 + 30*x3 + 65*x4 + 10*x5 + 35*x6\\nmodel.OF = Objective(rule=
objective_function, sense=maximize)\\nmodel.Constraint1 = Constraint(expr = 25*x1 + 50*x2 + 35*x3 + 45*x4 + 30*x5 + 20*x6 <= 8500)
\\nmodel.Constraint2 = Constraint(expr = x1 <= 14)\\nmodel.Constraint3 = Constraint(expr = x2 <= 14)\\nmodel.Constraint4 = 
Constraint(expr = x3 <= 14)\\nmodel.Constraint5 = Constraint(expr = x4 <= 14)\\nmodel.Constraint6 = Constraint(expr = x5 <= 14)\\n
model.Constraint7 = Constraint(expr = x6 <= 14)\\nmodel.Constraint8 = Constraint(expr = x4 <= 3*x2)\\nmodel.Constraint9 = 
Constraint(expr = x1 + x3 >= 4*y)\\nmodel.Constraint10 = Constraint(expr = x5 + x6 >= 6*(1 - y))
\end{verbatim}
} \\ 
\hline 
\multirow{3}{*}{\raisebox{-1.0\height}{\includegraphics[width=5cm]{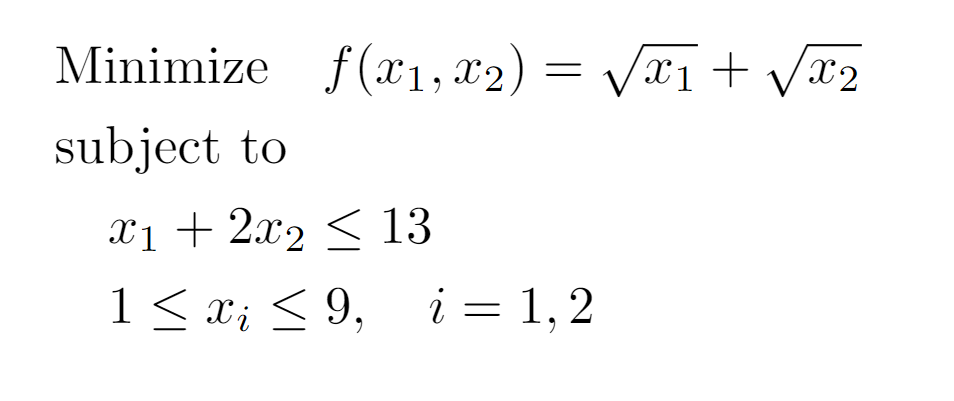}}} 
& \vtop{%

\begin{verbatim}
\text{Minimize} \quad f(x_1, x_2) = \sqrt{x_1} + \sqrt{x_2} \\ & \text{subject to} \\ & \quad x_1 + 2x_2 \leq 13 \\ & \quad 
1 \leq x_i \leq 9, \quad i = 1, 2
\end{verbatim}

} \\
\cline{2-2}
& \vtop{%
\begin{verbatim}
model.x1 = Var(bounds=(1,9))\nmodel.x2 = Var(bounds=(1,9))\ndef objective_function(model):\n    return sqrt(x1) + sqrt(x2)\n
model.obj = Objective(rule=objective_function, sense=minimize)\nmodel.Constraint1 = Constraint(expr = x1 + 2*x2 <= 13)
\end{verbatim}
} \\ 
\hline
\end{tabular}
\end{flushleft}
\end{table}

\end{landscape}
\newpage

\vspace{1cm}

\newpage

\section{Appendix: Module M1}\label{sec:Append_m1}
\vspace{-2mm}
We did 5 runs for AutoOpt-M1, Nougat, ChatGPT, and Gemini by randomly splitting the data with a training, validation, and test split of 80\%, 10\% and 10\%, respectively. The models corresponding to the median BLUE score for AutoOpt-M1, Nougat, ChatGPT and Gemini are reported in Section~\ref{sec:expt_m1}. The standard deviations in BLUE score and Character Error Rate are reported in Table~\ref{tab:std1} and Table~\ref{tab:std2}, respectively.

\begin{table}[htbp]
\caption{Standard deviation of BLUE score from 5 runs}
\centering
\begin{tabular}{lccc}
\toprule
\textbf{Model} & \textbf{HW} & \textbf{PR} & \textbf{HW+PR} \\
\midrule
GPT-4o              & 0.76          & 0.48          & 0.52           \\
Gemini 2.0 Flash    & 0.88          & 0.51          & 1.18           \\
Nougat              & 1.16          & 0.8          & 1.07           \\
AutoOpt-M1          & 1.14          & 0.87           & 1.04           \\
\bottomrule
\end{tabular}
\label{tab:std1}
\end{table}

\vspace{-3mm}
\begin{table}[htbp]
\caption{Standard deviation of Character Error Rate from 5 runs}
\centering
\begin{tabular}{lccc}
\toprule
\textbf{Model} & \textbf{HW} & \textbf{PR} & \textbf{HW+PR} \\
\midrule
GPT-4o              & 0.0068         & 0.0084          & 0.0032           \\
Gemini 2.0 Flash    & 0.0224         & 0.0054          & 0.0113           \\
Nougat              & 0.0098          & 0.0087          & 0.0058           \\
AutoOpt-M1          & 0.0122          & 0.0086          & 0.0072           \\
\bottomrule
\end{tabular}
\label{tab:std2}
\end{table}

\begin{figure}[h]
    \centering
    \includegraphics[scale=0.38]{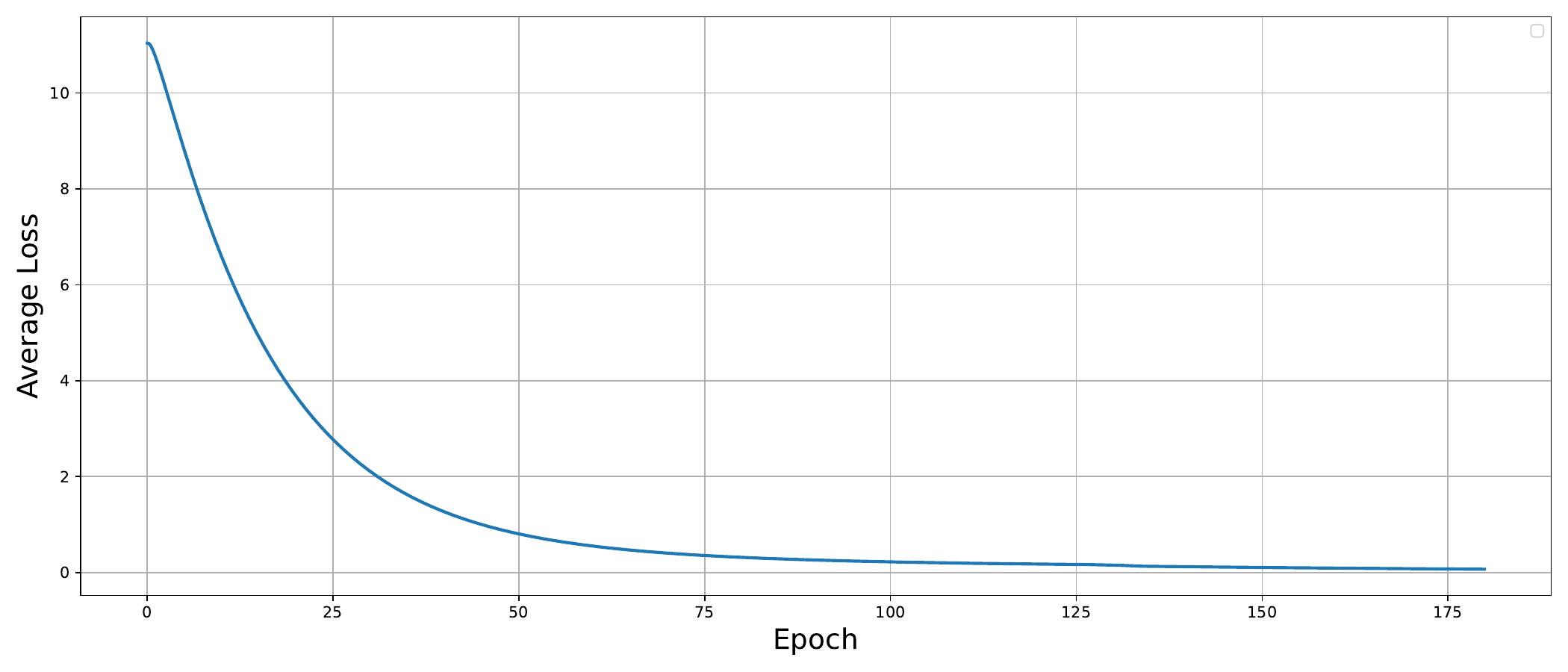}
    \caption{Convergence plot for AutoOpt-M1}
    \label{fig:convergencePlot}
\end{figure}

All experiments were conducted on Google Colab Pro using NVIDIA A100 GPU. The AutoOpt-M1 model was trained for 180 epochs. We used AdamW optimizer with learning rate 2$e^{-5}$, weight decay 0.02, and with a cosine scheduler. The batch size was set to 8 with gradient accumulation of 2. Each epoch took approximately 15-20 minutes. Figure~\ref{fig:convergencePlot} shows the convergence plot for the model for a particular run.

\vspace{-3mm}
\section{Appendix: Module M2}\label{sec:Append_m2}
\vspace{-2mm}
The median BLEU score and median Character Error Rate (CER) correspond to the same run among the five runs we performed. The median scores are reported in Section~\ref{sec:Module_M2}. The standard deviations for BLUE score (in percent) and Character Error Rate over 5 runs are 1.08 and 0.0051, respectively.

We fine-tuned the model for 15 epochs on Google Colab Pro using an NVIDIA A100 GPU with mixed-precision (fp16). For training, the batch size was set to 2 with gradient accumulation of 4 (effective batch size 8). The learning rate was set to 5$e^{-5}$ with a weight decay of 0.01. Each run took approximately 30–35 minutes.

\section{Appendix: Module M3}\label{sec:Append_m3}
\vspace{-2mm}
In this study, we consider a Bilevel Optimization based Decomposition (BOBD) method \cite{sinha2024decomposition} for optimization task in module M3 (Section \ref{sec:Module_M3}). BOBD method offers the advantages of both exact and approximation methods, by providing efficient solutions within a reasonable time frame. BOBD method attains this advantage by using a bilevel optimization structure that allows it to simultaneously use exact and approximation methods for problem solving. To explain the working procedure of the BOBD method, we first review the structures of general and bilevel optimization problems based on their formal definitions.

\begin{definition}\label{def:singleLevel}
A general optimization problem can be represented using its basic elements, decision variables $x=(x_1,\ldots,x_n)$, objective function $F(x)$, and constraints $G(x)$ and $H(x)$, as follows:
\begin{align}
\min_{x} \quad & F(x) \label{eq:startSingle} \\
\text{subject to} \quad & G_i(x)\leq 0, \quad i=1,\dots,I\\
& H_j(x) = 0, \quad j=1,\dots,J \label{eq:endSingle}
\end{align}
\end{definition}

Bilevel optimization problem is characterized by a unique structure in which the primary or upper level optimization problem contains an additional optimization problem, lower level optimization problem, nested within it as a constraint \cite{ViCa94, my-emo15, sinha2017review}. 
\begin{definition}\label{def:BiLevel}
A bilevel optimization problem, with upper level and lower level decision variables ($u$ and $l$), objective functions ($F(u,l)$ and $f(u,l)$), and constraints ($G(u,l)$ \& $H(u,l)$ and $g(u,l)$ \& $h(u,l)$), can be represented as follows:
\begin{align}
\min_{u,l} \quad & F(u,l) \label{eq:startBiLevel}\\
\text{subject to} &\notag\\  & \hspace{-12mm}l\in \argmin_{l} %
	\lbrace
		f(u,l) : g_p(u,l)\leq 0, \quad p=1,\dots,P, \notag\\ & \hspace{18mm} h_q(u,l) = 0, \quad q=1,\dots,Q
	\rbrace\\
& \hspace{-12mm}G_i(u,l)\leq 0, \quad i=1,\dots,I\\
 & \hspace{-12mm}H_j(u,l) = 0, \quad j=1,\dots,J \label{eq:endBiLevel}
\end{align}
\end{definition}

To solve a bilevel problem (Definition \ref{def:BiLevel}) in a nested manner, the values of upper level variables $u$ are fixed, and the lower level problem is solved with respect to $l$. By intelligently sampling $u$ and solving the lower level problem repeatedly, it is possible to converge to the bilevel optimum.

BOBD method involves the variable classification task, in which we classify each decision variable ($x_i \in x; \; i=1...n$) into upper level ($u$) or lower level ($l$) variables category. It allows to express the general optimization problem (Definition \ref{def:singleLevel}) in the form of bilevel optimization problem (Definition \ref{def:BiLevel}), which is called a bilevel decomposition process. In this study, we develop a Logistic Regression based Variable Classification Model (LR-VCM), a method to perform the variable classification task. To begin with, a population is generated and all the variables are initialized randomly. Thereafter, we start with a random approach of variable classification, where each decision variable is classified randomly into upper level (tag 0) or lower level (tag 1). Every single trial of level selection for all variables is referred to as Level Configuration (LC), and the collection of corresponding tag values constitutes a single observation for logistic regression. For given LC, we evaluate if it leads to an improvement in the objective function, when the lower level problem is solved. The improvement results are recorded on a binary scale: 0 indicates little improvement, and 1 indicates notable improvement. This binary score (0 or 1) acts as a label for a given LC. A pair of LC (observation) and corresponding improvement score (label) creates a single training sample for LR-VCM. We construct a training dataset by repeating the same procedure for the entire population. We perform a logistic regression using the developed training set and classify variables with statistically significant \textit{p}-values into lower level and the remaining variables into upper level. After variable classification ($u,l$), we intelligently sample the values of upper level variables $u$ using a genetic algorithm, and for each sample, the corresponding lower level problem is solved using a suitable classical optimization method, which provides the values of lower level variables $l$. The obtained values of upper level and lower level variables yield a particular solution for a given bilevel problem. This procedure is repeated several times to generate multiple solutions and an efficient solution $(u^*,l^*)$ is then identified from the generated solutions, as outlined in the pseudocode provided in Algorithm \ref{algo:BOBD_algo}.

\captionsetup[algorithm]{labelfont=bf,textfont=normalfont}
\begin{algorithm}
\caption{Bilevel Optimization-based Decomposition (BOBD)}\label{algo:BOBD_algo}
\begin{tabular}{cp{11.8cm}}
\textbf{Input}:  & $F(x)$, $G(x)$, $H(x)$: single level optimization problem (i.e., original problem)\\
\textbf{Output}: & $x^*=(u^*,l^*)$: the best solution found for single level optimization problem\vspace{0.5mm} \\
\hline \vspace{-3mm}\\
\textbf{Step 1}: \,  & Generate a population ($\P$) of random initial solutions.\\
\textbf{Step 2}: \,  & Develop a Logistic Regression based Variable Classification Model (LR-VCM).\\
\textbf{Step 3}: \,  & Perform a bilevel decomposition of original problem into upper level and lower level using LR-VCM.\\
\textbf{Step 4}: \,  & \textit{for} $g$ = 1 to \textit{$number\_of\_generations$}:\\
\textbf{Step 5}: \,  & \quad {If $g$ is divisible by \textit{$variable\_classification\_alternation\_number$} ($C$=10)}:\\
\textbf{Step 6}: \,  & \quad \quad \quad Develop a new LR-VCM using updated dataset.\\
\textbf{Step 7}: \,  & \quad \quad \quad Perform a new bilevel decomposition of original problem.\\
\textbf{Step 8}: \,  & \quad \ Sample the values of upper level variables $u$ using genetic algorithm.\\
\textbf{Step 9}: \,  & \quad \ For a given $u$, obtain $l$ by solving the corresponding lower level problem\\
       & \quad \ using the interior point or the linear programming methods.\\
\textbf{Step 10}: \,  & \quad \ Update population $\P$ with new solutions if they are better than the worst\\
& \quad \ solutions in the population.\\
\end{tabular}
\end{algorithm}

We bring novelty to BOBD method by incorporating LR-VCM for variable classification task. To evaluate the performance of the BOBD method, we consider a test suite of 10 optimization Test Problems (TP), TP1-TP10 \cite{sinha2024decomposition}. 
These test problems are derived from the real-world applications and exhibit various types of complexities such as non-convexity, non-linearity, non-differentiability, discreteness, high-dimensionality, etc. The description of test problems (TP1-TP10) and computational experiments are discussed next.

\vspace{3mm}
\textbf{TP1} (Structural Sensitivity Problem in a Chemical System \cite{stephanopoulos1975use}):
\begin{align*}
& \hspace{-45mm}\min_{x} \quad F(x)=x_1^{0.6}+x_2^{0.6}+x_3^{0.4}-4 x_3+2 x_4+5 x_5-x_6 \\
& \hspace{-42mm} \text{s.t.} \quad x_2-3 x_1-3 x_4=0;\\
& \hspace{-34mm} x_3-2 x_2-2 x_5=0;\\
& \hspace{-34mm} 4 x_4-x_6=0;\\
& \hspace{-34mm} x_1+2 x_4 \leq 4;\\
& \hspace{-34mm} x_2+x_5 \leq 4;\\
& \hspace{-34mm} x_3+x_6 \leq 6;\\
& \hspace{-34mm} x_1 \leq 3;\, x_3 \leq 4;\, x_5 \leq 2; \, x_1, x_2, x_3, x_4, x_5, x_6 \geq 0 
\end{align*}

\textbf{TP2} (Heat Exchanger Design Problem \cite{Liang2006ProblemDA}):
\begin{align*}
& \hspace{-23mm} \min_{x} \quad F(x)=x_1 \\
& \hspace{-19mm} \text{s.t.} \quad 35 x_2^{0.6}+35 x_3^{0.6}-x_1 \leq 0 ; \\
& \hspace{-15mm} -300 x_3+7500 x_5-7500 x_6-25 x_4 x_5+25 x_4 x_6+x_3 x_4=0 ; \\
& \hspace{-10mm} 100 x_2+155.365 x_4+2500 x_7-x_2 x_4-25 x_4 x_7-15536.5=0 ; \\
& \hspace{-14mm} -x_5+\ln \left(-x_4+900\right)=0 ; \\
& \hspace{-14mm} -x_6+\ln \left(x_4+300\right)=0 ; \\
& \hspace{-14mm} -x_7+\ln \left(-2 x_4+700\right)=0 ; \\
& \hspace{-10mm} 0 \leq x_1 \leq 1000 ;\quad 0 \leq x_2, x_3 \leq 40 ;\quad 100 \leq x_4 \leq 300 ;\\
& \hspace{-10mm} 6.3 \leq x_5 \leq 6.7 ;\quad 5.9 \leq x_6 \leq 6.4 ;\quad 4.5 \leq x_7 \leq 6.25
\end{align*}

\textbf{TP3} (More complexities added to TP1 \cite{sinha2024decomposition}):
\begin{align*}
& \hspace{-13mm} \min_{x} \quad F(x)= x_1^{0.6}+x_2^{0.6}+x_3^{0.4}-4 x_3+2 x_4+5 x_5-x_6+\frac{x_3^2}{16}-2 \cos \left(2 \pi x_2\right)\\
& \hspace{-10mm} \text{s.t.} \quad x_2-3 x_1-3 x_4=0; \\
&  \hspace{-3mm} x_3-2 x_2-2 x_5=0; \\
&  \hspace{-3mm} 4 x_4-x_6=0; \\
&  \hspace{-3mm}  x_1+2 x_4 \leq 4; \\
&  \hspace{-3mm} x_2+x_5 \leq 4; \\
&  \hspace{-3mm} x_3+x_6 \leq 6;\\
&  \hspace{-3mm} x_1 \leq 3 ;\quad x_5 \leq 2 ;\quad x_3 \leq 4;\\
&  \hspace{-3mm} x_1, x_2, x_3, x_4, x_5, x_6 \geq 0 
\end{align*}
\vspace{3mm}

\textbf{TP4} (Scalable variables $y$ and constraints added to TP3 \cite{sinha2024decomposition}):
\begin{align*}
& \hspace{-25mm} \min_{x} \quad F(x) = x_1^{0.6} + x_2^{0.6} + x_3^{0.4} - 4 x_3 + 2 x_4 + 5 x_5 - x_6 + \frac{x_3^2}{16} - \frac{x_2^2}{16}\\
& \hspace{-3mm} - 2 \cos(2 \pi x_3) - 2 \cos(2 \pi x_2) + \sum_{p=1}^P y_p \cdot x_1^{0.6} \\
& \hspace{-22mm} \text{s.t.} \quad x_2 - 3 x_1 - 3 x_4 = 0; \\
& \hspace{-14mm} x_3 - 2 x_2 - 2 x_5 = 0; \\
& \hspace{-14mm} x_1 + 2 x_4 \leq 4; \\
& \hspace{-14mm} x_2 + x_5 \leq 4; \\
& \hspace{-14mm} x_3 + x_6 \leq 6; \\
& \hspace{-14mm} \sqrt{x_1 + x_2 + x_3} - y_p \leq 0, \quad \forall p; \\
& \hspace{-14mm} x_1 \leq 3; \quad x_5 \leq 2; \quad x_3 \leq 4; \\
& \hspace{-14mm} 1 \leq y_p \leq 5, \, \forall p; \\
& \hspace{-14mm} x_1, x_2, x_3, x_4, x_5, x_6 \geq 0
\end{align*}
\vspace{3mm}

\textbf{TP5} (Scalable variables $y$ and constraints added to TP2 \cite{sinha2024decomposition}):
\begin{align*}
& \hspace{-14mm} \min_{x} \quad F(x)= x_1-50 \cos \left(2 \pi x_4\right)+\sum_{p=1}^P \frac{y_p{ }^2}{x_4} \\
& \hspace{-12mm} \text{s.t.} \quad 35 x_2^{0.6}+35 x_3^{0.6}-x_1 \leq 0 ; \\
& \hspace{-9mm} -300 x_3+7500 x_5-7500 x_6-25 x_4 x_5+25 x_4 x_6+x_3 x_4=0 ; \\
& \hspace{-5mm} 100 x_2+155.365 x_4+2500 x_7-x_2 x_4-25 x_4 x_7-15536.5=0 ; \\
& \hspace{-9mm} -x_5+\ln \left(-x_4+900\right)=0 ; \\
& \hspace{-9mm} -x_6+\ln \left(x_4+300\right)=0 ; \\
& \hspace{-9mm} -x_7+\ln \left(-2 x_4+700\right)=0 ; \\
& \hspace{-5mm} x_4^{0.2}+x_5+x_6-y_p \leq 0, \hspace{1.5mm}\forall p ; \\
& \hspace{-5mm} 0 \leq x_1 \leq 1000 ;\quad 0 \leq x_2, x_3 \leq 40 ;\quad 100 \leq x_4 \leq 300 ;\\
& \hspace{-5mm} 6.3 \leq x_5 \leq 6.7 ;\quad 5.9 \leq x_6 \leq 6.4 ;\quad 4.5 \leq x_7 \leq 6.25; \quad 10 \leq y_p \leq 30, \hspace{0.5mm} \forall p
\end{align*}
\vspace{3mm}
\newpage

\textbf{TP6} (Scalable variables ($y,z$) and constraints added to existing problem \cite{floudas1990collection}):
\begin{align*}
& \hspace{3mm} \min_{x} \quad F(x) = -25\left(x_1-2\right)^2-\left(x_2-2\right)^2-\left(x_3-1\right)^2-\left(x_4-4\right)^2-\left(x_5-1\right)^2-\left(x_6-4\right)^2\\
& \hspace{25mm} +\sum_{p=1}^P\left(x_3-y_p\right)^2-\sum_{\mathrm{q}=1}^Q\left(x_5-z_q\right)^2\\
& \hspace{5mm} \text{s.t.} \quad -\left(x_3-3\right)^2-x_4+4 \leq 0;\\
& \hspace{13mm} -\left(x_5-3\right)^2-x_6+4 \leq 0; \\
& \hspace{13mm} -x_1-x_2+2 \leq 0; \\
& \hspace{18mm} x_1-3 x_2 \leq 2 ;\\ 
& \hspace{18mm} x_2-x_1 \leq 2 ; \\
& \hspace{18mm} x_1+x_2 \leq 6; \\
& \hspace{18mm} y_p-x_3+1 \leq 0,\hspace{1.5mm} \forall p ;\\
& \hspace{18mm} z_q^2-x_3^2-x_5^2 \leq 0,\hspace{1.5mm} \forall q;\\
& \hspace{18mm} 0 \leq x_1; \quad 0 \leq x_2 ;\quad 1 \leq x_3 \leq 5 ;\quad 0 \leq x_4 \leq 6;\\
& \hspace{18mm} 1\leq x_5 \leq 5 ;\quad 0 \leq x_6 \leq 10 ;\quad 0 \leq y_p \leq 5, \hspace{0.5mm} \forall p;\quad 0 \leq z_q \leq 5, \hspace{0.5mm} \forall q
\end{align*}

\vspace{5mm}
\textbf{TP7} (Pool blending Problem with additional complexities \cite{Liang2006ProblemDA}):
\begin{align*}
& \hspace{3mm} \min_{x} \quad F(x) = 6 x_1+16 x_2-9 x_5+10\left(x_6+x_7\right)-15 x_8+x_9^2+50 \cos \left(\pi x_9\right)-25 \cos \left(\pi x_8\right)\\
& \hspace{25mm} -\ln \left(x_8-x_9\right)-\sum_{p=1}^P\left(y_p-x_9\right)^2+\sum_{q=1}^Q\left(z_q-x_8\right)^2 \\
& \hspace{5mm} \text{s.t.} \quad x_1+x_2-x_3-x_4=0 ;\\
& \hspace{13mm} x_3+x_6-x_5=0 ;\\
& \hspace{13mm} x_4+x_7-x_8=0 ; \\
& \hspace{13mm} 0.03 x_1+0.01 x_2-x_3 x_9-x_4 x_9=0 ; \\
& \hspace{13mm} x_3 x_9+0.02 x_6-0.025 x_5 \leq 0 ; \\
& \hspace{13mm} x_4 x_9+0.02 x_7-0.015 x_8 \leq 0 ; \\
& \hspace{13mm} x_9^2-y_p^2 \leq 0,\hspace{1.5mm} \forall p ; \\
& \hspace{13mm} x_8^2-z_q^2 \leq 0,\hspace{1.5mm} \forall q ; \\
& \hspace{13mm} 0 \leq x_1, x_2, x_6 \leq 300 ;\quad 0 \leq x_3, x_5, x_7 \leq 100 ;\quad 0 \leq x_4, x_8 \leq 200;\\
& \hspace{13mm} 0.01 \leq x_9 \leq 0.03 ;\quad 0 \leq y_p \leq 1, \hspace{0.5mm} \forall p;\quad 1 \leq z_q \leq 200, \hspace{0.5mm} \forall q
\end{align*}
\vspace{5mm}

\textbf{TP8} (Existing problem with additional complexities \cite{sinha2024decomposition}):
\begin{align*}
& \hspace{-5mm} \min_{x} \quad F(x) = 5 \sum_{i=1}^4 x_i-5 \sum_{i=1}^4 x_i^2-\sum_{i=5}^{13} x_i-20 e^{-0.1 \sqrt{\sum_{i=1}^4 x_i^2}}-e^{0.25 \sum_{\mathrm{i}=1}^4 \cos \left(2 \pi x_i\right)} \\
& \hspace{16mm} +\sum_{p=1}^P\left(y_p^2+\sum_{\mathrm{i}=1}^4 \cos \left(2 \pi x_i\right)\right) \\
& \hspace{-3mm} \text{s.t.} \quad 2 x_1+2 x_2+x_{10}+x_{11} \leq 10; \\
& \hspace{5mm} 2 x_1+2 x_3+x_{10}+x_{12} \leq 10 ; \\
& \hspace{5mm} 2 x_2+2 x_3+x_{11}+x_{12} \leq 10 ; 
\end{align*}
\newpage

\begin{align*}
& \hspace{-12mm}  x_{10}-8 x_1 \leq 0 ; \\
& \hspace{-12mm}  x_{11}-8 x_2 \leq 0 ; \\
& \hspace{-12mm}  x_{12}-8 x_3 \leq 0 ; \\
& \hspace{-12mm}  x_{10}-x_5-2 x_4 \leq 0 ; \\
& \hspace{-12mm}  x_{11}-x_7-2 x_6 \leq 0 ; \\
& \hspace{-12mm}  x_{12}-x_9-2 x_8 \leq 0 ; \\
& \hspace{-12mm} \sum_{\mathrm{i}=1}^4 \cos \left(2 \pi x_i\right)-y_p \leq 0, \hspace{1mm} \forall p ; \\
& \hspace{-12mm}  0 \leq x_i \leq 3 \hspace{1.5mm}(i=1, \ldots, 4); \quad 0 \leq x_i \leq 1 \hspace{1.5mm}(i=5, \ldots, 9);\\
&  \hspace{-12mm} 0 \leq x_i \leq 100 \hspace{1.5mm}(i=10,11,12); \quad 0 \leq x_{13} \leq 1 ; \quad -5 \leq y_p \leq 5, \hspace{0.5mm} \forall p\\
\end{align*}

\textbf{TP9} (Heat Exchanger Design Problem with additional complexities \cite{Liang2006ProblemDA}):
\begin{align*}
& \hspace{-23mm} \min_{x} \quad F(x) = x_1+x_2+x_3+\sum_{p=1}^P\left(\tan \left(y_p\right)-15 \cos 2 \pi\left(x_1+x_2+x_3\right)\right)^2\\
& \hspace{-20mm} \text{s.t.} \quad  0.0025 x_4+0.0025 x_6 \leq 1 ; \\
& \hspace{-12mm}  0.0025 x_5+0.0025 x_7-0.0025 x_4 \leq 1 ;\\
& \hspace{-12mm}  0.01 x_8-0.01 x_5 \leq 1 ;\\
& \hspace{-16mm} -x_1 x_6+100 x_1+833.33 x_4 \leq 83333.33; \\
& \hspace{-16mm} -x_2 x_7+1250 x_5-1250 x_4+x_2 x_4 \leq 0; \\
& \hspace{-16mm} -x_3 x_8-2500 x_5+x_3 x_5+1250000 \leq 0; \\
& \hspace{-12mm} \tan \left(y_p\right)-\ln \left(x_1+x_2+x_3\right) \leq 0 \quad \forall p;\\
& \hspace{-16mm} -\tan \left(y_p\right)-\ln \left(x_1+x_2+x_3\right) \leq 0 \quad \forall p; \\
& \hspace{-12mm} 100 \leq x_1 \leq 10000; \quad 1000 \leq x_i \leq 10000 \hspace{1.5mm} (i=2,3); \\
& \hspace{-12mm} 10 \leq x_i \leq 1000  \hspace{1.5mm}(i=4, \ldots, 8); \quad -\pi / 2 \leq  y_p \leq \pi / 2, \hspace{0.5mm} \forall p\\
\end{align*}

\textbf{TP10} (Existing problem with additional complexities \cite{sinha2024decomposition}):
\begin{align*}
& \hspace{-15mm} \min_{x} \quad F(x) = 37.293239x_1 + 0.8356891x_1x_5 + 5.3578547x_3^2 - 40792.14 \\
& \hspace{6mm} + \sum_{p=1}^{P}(y_p - x_1 - x_3)^2- 150\sum_{q=1}^{Q}\cos(2\pi z_q) \\
& \hspace{-11mm} \text{s.t.} \quad  0.0056858x_2x_5 - 0.0022053x_3x_5 + 0.0006262x_1x_4 \le 6.665593; \\
&  \hspace{-7mm} -0.0056858x_2x_5 + 0.0022053x_3x_5 - 0.0006262x_1x_4 \le 85.334407; \\
&  \hspace{-3mm} 0.0071317x_2x_5 + 0.0021813x_3^2 + 0.0029955x_1x_2 \le 29.48751; \\
& \hspace{-7mm} -0.0071317x_2x_5 - 0.0021813x_3^2 - 0.0029955x_1x_2 + 9.48751 \le 0; \\
& \hspace{-3mm} 0.0047026x_3x_5 + 0.0019085x_3x_4 + 0.0012547x_1x_3 \le 15.699039; \\
& \hspace{-7mm} -0.0047026x_3x_5 - 0.0019085x_3x_4 - 0.0012547x_1x_3 + 10.699039 \le 0; \\
& \hspace{-3mm}  y_p - \ln(x_1 + x_3 + 1) \le 0, \hspace{1mm} \forall p; \\
& \hspace{-3mm}  z_q^3 - x_1^3 - x_3^3 - x_5^3 \le 0, \hspace{1mm} \forall q; \\
& \hspace{-3mm}  78 \le x_1 \le 102; \quad 33 \le x_2 \le 45; \quad 27 \le x_3, x_4, x_5 \le 45;\\
& \hspace{-3mm}  0 \le y_p \le 5, \hspace{0.5mm} \forall p; \quad -5 \le z_q \le 5, \hspace{0.5mm} \forall q \\
\end{align*}
\newpage

For computational experiments, we consider the following three scenarios: small-scale ($|y|+|z|=0$), medium-scale ($|y|+|z|=20$), and large-scale ($|y|+|z|=50$) (here, $|y|+|z|$ represents the number of scalable variables and constraints in TP). All test problems are solved in small to large scale scenarios using the Interior Point (IP), Genetic Algorithm (GA), and BOBD methods. For all TP, constraint tolerance is set to $10^{-4}$. 
For genetic algorithm, the details on GA operators and parameters value are provided in Table \ref{tab:GA_Param}. For GA implemented in BOBD method, all parameters are kept same as mentioned in Table \ref{tab:GA_Param}, and an improvement based-termination criterion is used. We consider a steady state genetic algorithm \cite{syswerda1991study} in both explicit GA and GA used in BOBD. 

\begin{table}[h]
\centering
\caption{Genetic algorithm operators and parameter values for computational experiments}
\begin{tabular}{|p{3.5cm}|p{5.8cm}|>{\centering\arraybackslash}m{3.3cm}|}
\hline
\textbf{GA operators} & \hspace{2cm} \textbf{Mechanisms} & \textbf{Parameters value} \\
\hline
Crossover & simulated binary crossover (SBX) \cite{deb1995simulated} & 0.90 \\
\hline
Mutation & polynomial mutation \cite{deb2014analysing} & 0.10 \\
\hline
Selection & tournament selection \cite{miller1995genetic} & -- \\
\hline
Population size & \hspace{2cm}--- & 200 \\
\hline
No. of offsprings & \hspace{2cm}--- & 2 \\
\hline
\end{tabular}
\label{tab:GA_Param}
\end{table}

\begin{figure*}[h]
\centering
\includegraphics[width=1.1\textwidth]{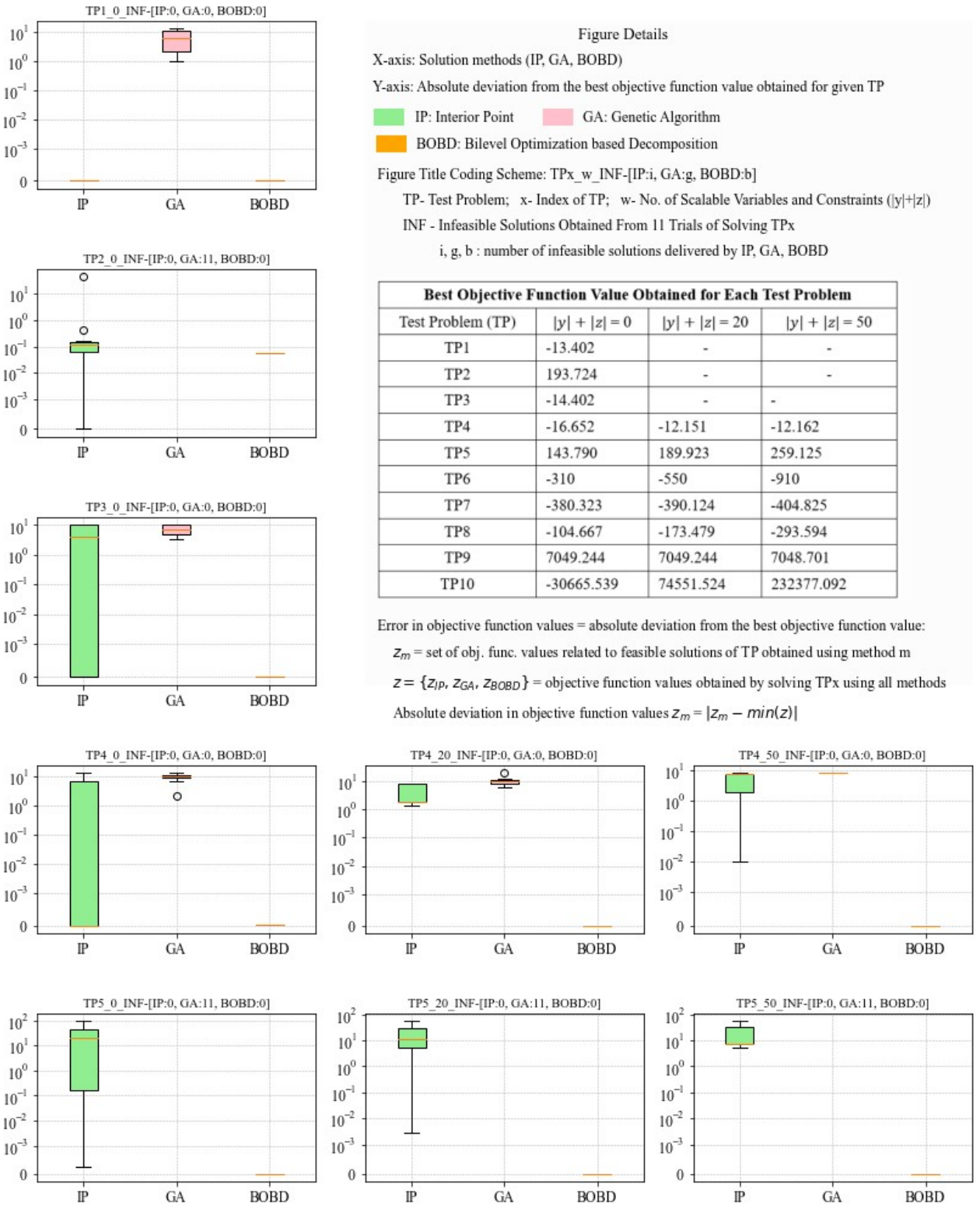}
\caption{Absolute deviation in objective function values from 11 runs: TP1-TP5}
\vspace{-3mm}
\label{fig:abs_dev_1}
\end{figure*}

\begin{figure*}[hbt]
\centering
\includegraphics[width=1.1\textwidth]{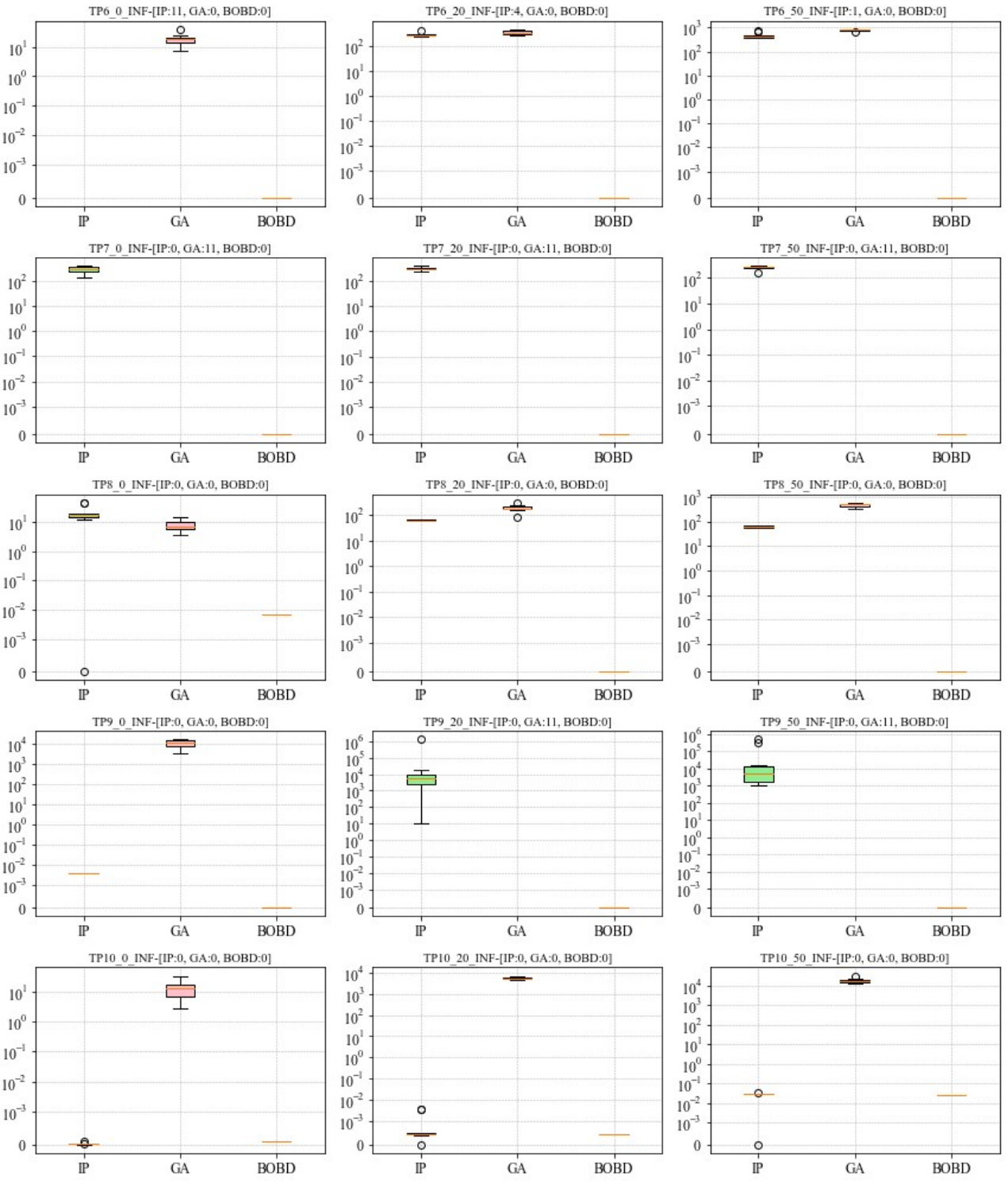}
\caption{Absolute deviation in objective function values from 11 runs: TP6-TP10}
\vspace{-3mm}
\label{fig:abs_dev_2}
\end{figure*}

Each test problem (TP1-TP10) is solved 11 times using the IP, GA, and BOBD methods and the corresponding objective function values are recorded. For each TP, the best feasible objective function value (obtained or known from the literature) is recorded. Every time a method is executed, we evaluate the quality of solution in terms of absolute deviation, which is the absolute difference in the solution obtained from the method and the best known solution.
These deviations are illustrated using box-plots in Figure~\ref{fig:abs_dev_1} and Figure~\ref{fig:abs_dev_2}. 
The figures indicate that BOBD method consistently yields the best solutions across all 11 runs for every test problem. In contrast, IP and GA frequently converge to local optima and, in several cases, even provide infeasible solutions. The BOBD method either matches or outperforms the solutions obtained by IP and GA in all instances. In the case of BOBD, the solutions are at least the same or better compared to the results obtained from IP and GA. For certain problems in figures, BOBD performance appears to be slightly worse than IP. This is because, in such cases, both BOBD and IP have converged close to the optimum and the difference in solution quality is marginal.

We also record the average computational time for solving all instances using BOBD. The IP method typically terminates within 1–10 seconds for most of the test problems. Computational times for BOBD method are provided in Table \ref{tab:compu_time}. For a fair comparison, GA is allowed to run for twice the time taken by the BOBD method for each instance. Overall, the data in Figure \ref{fig:abs_dev_1}, Figure \ref{fig:abs_dev_2}, and Table \ref{tab:compu_time} convey that IP method provides the solutions quickly but it often converges to suboptimal solutions. GA frequently struggles to find even a feasible solution, particularly in medium and large scale scenarios. BOBD method took more computational time than IP, but it consistently delivers high-quality solutions during all 11 runs, which demonstrates better accuracy and repeatability of BOBD compared to IP and GA methods.

\begin{table}[h]
\centering
\caption{Average computational time for BOBD method (in seconds)}
\begin{tabular}{|c|p{2.5cm}|p{2.5cm}|p{2.5cm}|}
\hline
\textbf{Test Problem} & \hspace{0.3cm} \textbf{$|y|+|z|=0$} & \hspace{0.3cm} \textbf{$|y|+|z|=20$} & \hspace{0.3cm} \textbf{$|y|+|z|=50$} \\
\hline
TP1 & 1.90 & - & - \\
TP2 & 4.00 & - & - \\
TP3 & 3.34 & - & - \\
TP4 & 3.86 & 28.62 & 32.49 \\
TP5 & 4.34 & 8.77 & 12.33 \\
TP6 & 7.72 & 8.26 & 21.00 \\
TP7 & 8.33 & 11.26 & 51.21 \\
TP8 & 16.16 & 4.83 & 5.79 \\
TP9 & 14.83 & 291.26 & 496.70 \\
TP10 & 27.23 & 28.67 & 32.42 \\
\hline
\end{tabular}
\label{tab:compu_time}
\end{table}

\end{document}